\documentclass[10pt,twocolumn,letterpaper]{article}

\usepackage{iccv}
\usepackage{times}
\usepackage{epsfig}
\usepackage{graphicx}
\usepackage{amsmath}
\usepackage{amssymb}
\usepackage{mathrsfs}
\usepackage{amsthm}
\usepackage{array}
\usepackage{multirow}
\usepackage{amsfonts}
\usepackage{booktabs}
\usepackage{subfigure}
\usepackage{float}
\usepackage{algorithm}
\usepackage{algpseudocode}
\usepackage{makecell}
\usepackage{caption}
\usepackage{url}

\usepackage[breaklinks=true,bookmarks=false]{hyperref}

\iccvfinalcopy 

\def\iccvPaperID{****} 

\ificcvfinal\fi

\begin{document}

\title{Online Multi-Granularity Distillation for GAN Compression}

\author{Yuxi Ren\footnotemark[1] 
\qquad  Jie Wu\footnotemark[1]   
 \qquad Xuefeng Xiao 
 \qquad Jianchao Yang\\
ByteDance Inc.\\
{\tt\small \{renyuxi.20190622, wujie.10, xiaoxuefeng.ailab, yangjianchao\}@bytedance.com}
}

\twocolumn[{%
\renewcommand\twocolumn[1][]{#1}%
\maketitle
\begin{center}
    \centering
    \vspace{-20pt}
    \includegraphics[width=0.99\linewidth]{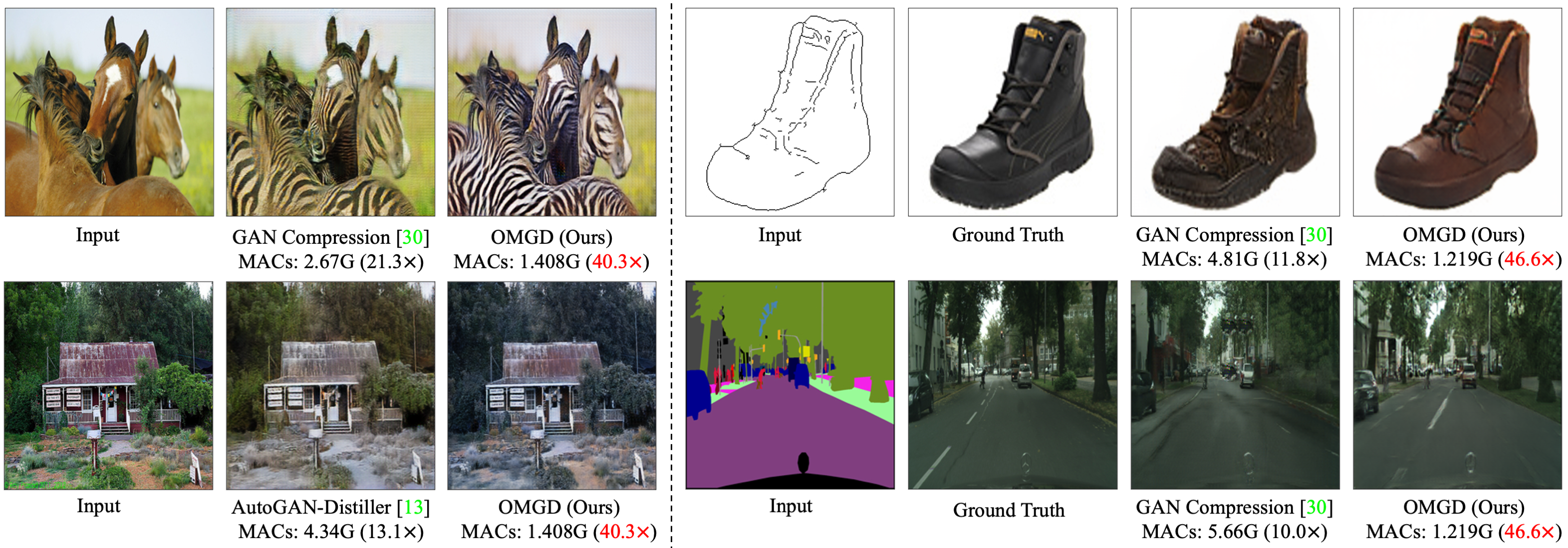}
    \vspace{-5pt}
    \captionof{figure}{We introduce Online Multi-Granularity Distillation (OMGD) scheme for compressing conditional GANs. OMGD reduces the computation of Pix2Pix and CycleGAN by \textbf{40.3-46.6$\times$} while preserving the visual fidelity.}
    \vspace{5pt}
    \label{fig:abstract}
\end{center}%
}]

\maketitle
\footnotetext[1]{These authors contributed equally to this work.}

\ificcvfinal\thispagestyle{empty}\fi

\begin{abstract}
Generative Adversarial Networks (GANs) have witnessed prevailing success in yielding outstanding images, however, they are burdensome to deploy on resource-constrained devices due to ponderous computational costs and hulking memory usage. 
Although recent efforts on compressing GANs have acquired remarkable results, they still exist potential model redundancies and can be further compressed.
To solve this issue, we propose a novel online multi-granularity distillation (OMGD) scheme to obtain lightweight GANs, which contributes to generating high-fidelity images with low computational demands.
We offer the first attempt to popularize single-stage online distillation for GAN-oriented compression, where the progressively promoted teacher generator helps to refine the discriminator-free based student generator.
Complementary teacher generators and network layers provide comprehensive and multi-granularity concepts to enhance visual fidelity from diverse dimensions.
Experimental results on four benchmark datasets demonstrate that OMGD successes to compress  \textbf{40$\times$} MACs and \textbf{82.5$\times$} parameters on Pix2Pix and CycleGAN, without loss of image quality. 
It reveals that OMGD provides a feasible solution for the deployment of real-time image translation on resource-constrained devices. Our code and models are made public at: \url{https://github.com/bytedance/OMGD}
\end{abstract}

\begin{figure*}
    \centering
    \includegraphics[width=0.99\linewidth]{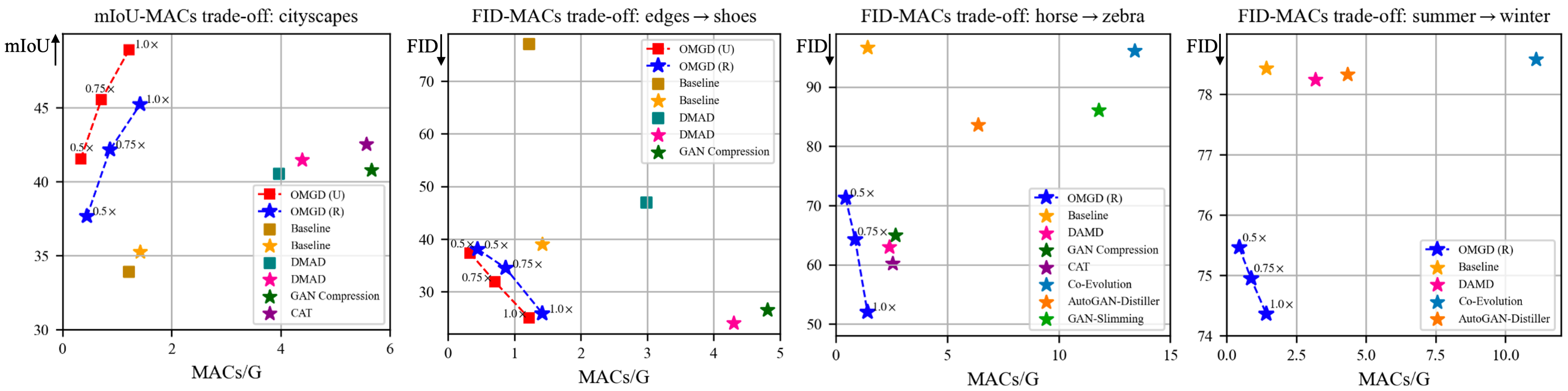}
    \vspace{-5pt}
    \captionof{figure}{Performance-MACs trade-off between OMGD and existing competitive methods including GAN compression \cite{GAN_compression}, CAT \cite{CAT}, DMAD \cite{DMAD}, GAN-Slimming \cite{GAN_slimming}, AutoGAN-Distiller \cite{AGD} and Co-Evolution \cite{co_evolution}. $\square$ denotes the U-Net style generator and $\star$ is the Res-Net style. OMGD significantly outperforms these methods with much less computational costs.
    “Baseline" denotes that the model is trained with naive GAN loss. 
    }
    \label{fig:performance}
    \vspace{-5pt}
\end{figure*}%

\section{Introduction}
Recently, Generative Adversarial Networks (GANs) \cite{gan} have achieved prominent results in diversified visual applications, such as image synthesis \cite{DCGAN, Reed2016GenerativeAT, SNGAN, Self-Attention-GAN, bigGANs} and image-to-image translation \cite{pix2pix,cyclegan,CartoonGAN, StyleGAN, StarGAN, Tang_2019_CVPR}.
Albeit with varying degrees of progress, most of its recent successes \cite{pix2pix,cyclegan,Tang_2019_CVPR, Self-Attention-GAN, CartoonGAN, bigGANs} are involved in huge resource demands.
It is arduous to popularize such models that require tremendous computational costs, which becomes a critical bottleneck as this model is deployed on resource-constrained mobile phones or other lightweight IoT devices \cite{MobileNet_V1, HAQ, GAN_compression,portableGAN}. 
To alleviate such expensive and unwieldy computational costs, GAN compression becomes a newly-raised and crucial task.
A great deal of mainstream model compression techniques 
\cite{li2019oicsr,li2020pams,xiao2017design,qi2021learning,xiao2017building,lin2021network} are employed to learn efficient GAN, including knowledge distillation \cite{GAN_compression,GAN_KD,GAN_slimming, TinyGAN,AGD,AutoGAN, DMAD, slimmable_GAN, portableGAN,CAT}, channel pruning \cite{GAN_compression,DMAD,GAN_slimming} and nerual architecture search \cite{AutoGAN,GAN_compression, DMAD}.

However, the above compression algorithms primarily exist threefold issues.
Firstly, they tend to straightforwardly resort to the mature model compression techniques \cite{OFA, SNNs, KD}, which are not customized for GAN and lack exploration of complex characteristics and structures for GAN.
Secondly, they usually formulate GAN compression as a multi-stages task. For example, \cite{GAN_compression} needs to pre-train, distillate, evolute, and fine-tune sequentially. The distillation-based methods \cite{GAN_compression,DMAD,AGD,portableGAN,TinyGAN,GAN_slimming,CAT} should pre-train a teacher generator and then distill the student one. 
An end-to-end approach is essential to reduce the complicated time and computational resources in the multi-stages setting.
Thirdly, the current state-of-the-art methods are still burdened with high computational costs. For instance, the best model \cite{DMAD} requires 3G MACs, which is relatively high for deployment on lightweight edge devices.
 
To overcome the above issues, we craft to propose a novel \textit{Online Multi-Granularity Distillation (OMGD)} framework for learning efficient GANs.
We abandon the complex multi-stage compression process and design a GAN-oriented online distillation strategy to obtain the compressed model in one step. We can excavate potential complementary information from multiple levels and granularities to assist in the optimization of compressed models. These concepts can be regarded as auxiliary supervision cues, which is very critical to break through the bottleneck of capacity for models with low computational costs.
The contributions of OMGD can be summarized as follows:
\begin{itemize}
\item 
To the best of our knowledge, we offer the first attempt to popularize distillation to an online scheme in the field of GAN compression and optimize the student generator in a discriminator-free and ground-truth-free setting.
This scheme trains the teacher and student alternatively, promoting these two generators iteratively and progressively.
The progressively optimized teacher generator helps to warm up the student and guide the optimization direction step by step.

\item We further extend the online distillation strategy into a multi-granularity scheme from two perspectives.
On the one hand, we employ different structure based teacher generators to capture more complementary cues and enhance visual fidelity from more diversified dimensions.
On the other hand, besides the concepts of the output layer, we also transfer the channel-wise granularity information from intermediate layers to play as additional supervisory signals.

\item  Extensive experiments on widely-used datasets (i.e., horse$\rightarrow$zebra \cite{cyclegan}, summer$\rightarrow$winter \cite{cyclegan}, edges$\rightarrow$shoes \cite{Yu2014FineGrainedVC} and cityscapes \cite{Cordts2016TheCD}) demonstrate that OMGD can reduce the computation of two essential conditional GAN models including pix2pix \cite{pix2pix} and CycleGAN \cite{cyclegan} by \textbf{40$\times$} regarding MACs, without loss of the visual fidelity of generated images.
It reveal that OMGD is efficient and robust for various benchmark datasets, diverse conditional GAN, network architectures as well as problem settings (paired or unpaired).
Compared with the existing competitive approaches, OMGD helps to obtain better image quality with much less computational costs (see Figure \ref{fig:abstract} and \ref{fig:performance}). 
Furthermore, OMGD 0.5$\times$ (only requires \textbf{0.333G MACs}) successes to achieve impressive results, which provides a feasible solution for deployment on resource-constrained devices and even breaks the barriers to real-time image translation on mobile devices.
\end{itemize}

\section{Related Work}

\begin{figure*}[ht]
   \centering
   \includegraphics[width=17cm,height=8.0cm]{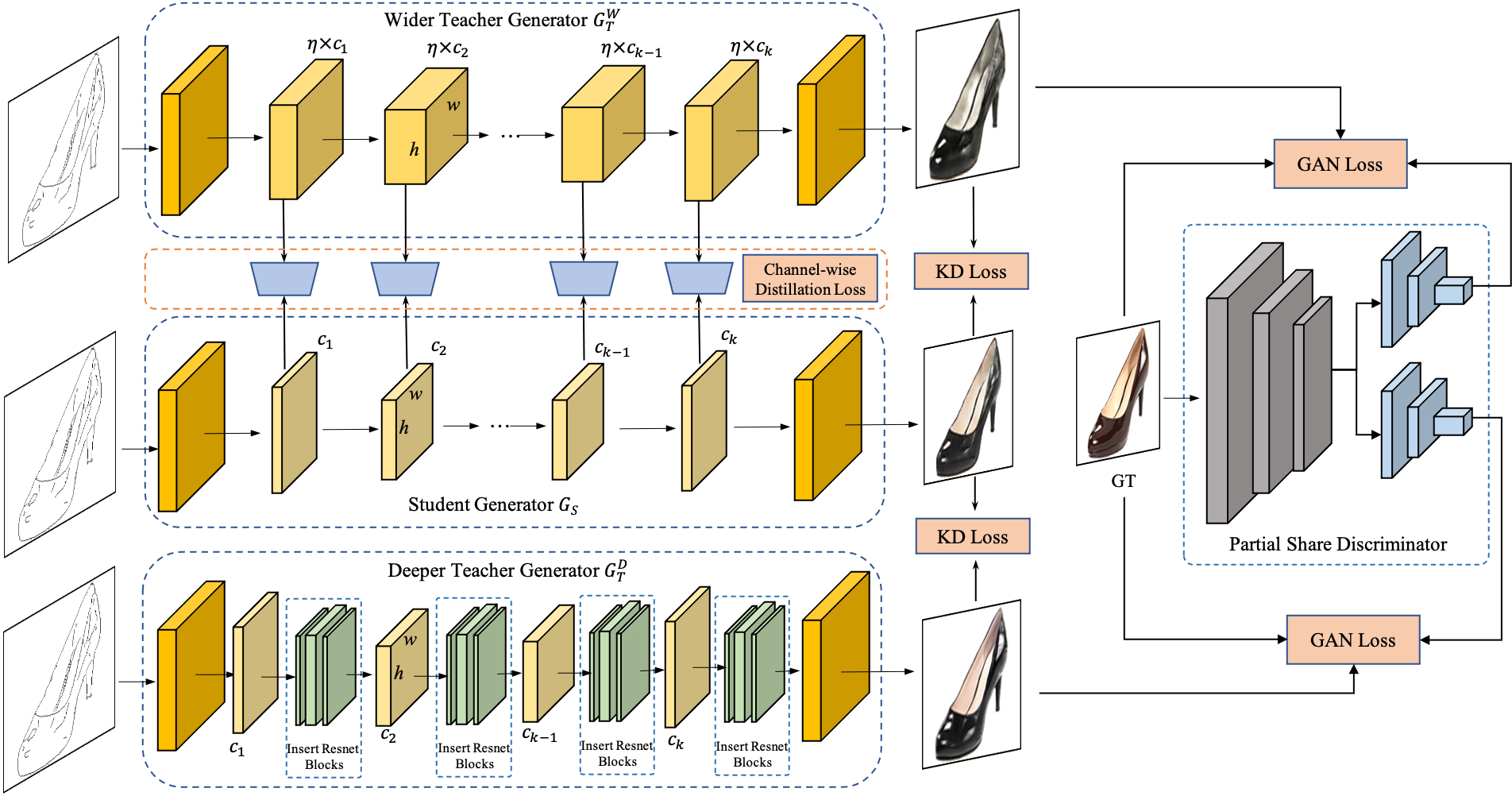}
   \vspace{-2pt}
   \caption{The pipeline of Online Multi-Granularity Distillation framework. 
   The student generator $G_S$ only leverages the complementary teacher generators ($G_T^W$ and $G_T^D$) for optimization and can be trained in the discriminator-free and ground-truth-free setting.
   This framework transfers different levels concepts from the intermediate layers and output layer to perform the knowledge distillation. The whole optimization is conducted on a online distillation scheme. Namely, $G_{T}^W$, $G_{T}^D$ and $G_{S}$ are optimized simultaneously and progressively.
   }
   \vspace{-5pt}
   \label{fig:MGOKD}
   \end{figure*}

\subsection{GANs and GAN Compression} 
Generative Adversarial Networks (GANs) \cite{gan} have obtained impressive results on a series of computer vision tasks, such as image-to-image translation \cite{cyclegan,pix2pix, CartoonGAN, StyleGAN, StarGAN, Tang_2019_CVPR, GauGAN}, image generation \cite{DCGAN, Reed2016GenerativeAT, SNGAN, Self-Attention-GAN, bigGANs, imagegeneration} and image inpainting \cite{Pathak_2016_CVPR, Yang_2017_CVPR, Wan2020BringingOP, imagepainting}.
Specifically, Pix2Pix \cite{pix2pix} conducted a min-max game between a generator and a discriminator to employ paired training data for image-to-image translation. 
CycleGAN \cite{cyclegan} further extended the capacity of GANs for image translation in a weakly-supervised setting,
where no paired data are leveraged in the training stage.
Although various GAN methods have achieved impressive successes recently, they tend to occupy a growing number of memory and computational costs \cite{CartoonGAN, GauGAN, bigGANs} to support their powerful performance, which conflicts the deployment on resource-constrained devices.

Recently, GAN-oriented compression has become an essential task due to its potential applications in the field of lightweight device deployment.
Shu \etal \cite{co_evolution} presented the first preliminary study on introducing the co-evolutionary algorithm for removing redundant filters to compress CycleGAN.
Fu \etal \cite{AGD} employed AutoML approaches and searched an efficient generator with the guide of the target computational resource constraints. 
Wang \etal \cite{GAN_slimming} proposed a unified GAN compression optimization framework including model distillation, channel pruning and quantization. 
Li \etal \cite{GAN_compression} designed a "once-for-all" generator which decouples the model training and architecture search via weight sharing.
Li \etal \cite{DMAD} proposed a differentiable mask and a co-attention distillion algorithm to learn effective GAN.
Jin \etal \cite{CAT} proposed a one-step pruning algorithm to search a student model from the teacher model.
In this work, we design an Online Multi-Granularity Distillation (OMGD) scheme. By introducing multi-granularity knowledge guidance, the student generator can be enhanced by leveraging the complementary concepts from diverse teachers and layers, which intrinsically improves the capacity of the compressed model. 

\subsection{Knowledge Distillation} 
Knowledge Distillation (KD) \cite{KD} is a fundamental compression technique, where a smaller student model is optimized under the effective information transfer and supervision of a larger teacher model or ensembles \cite{model_compression}. Hinton \cite{KD} performed knowledge distillation via minimizing the distance between the output distribution statistics between student and teacher network. 
In this way, the student network attempts to learn dark knowledge \cite{KD} that contains the similarities between different classes, which can not be provided by the ground truth labels. 
Romero \etal \cite{FitNets} further took advantage of the concepts of feature maps from the intermediate layers to enhance the performance of the student network. 
Zhou \etal \cite{channel_KD} presented that each channel of the feature map corresponds to a visual pattern, so they focus on transferring attention concepts \cite{wu2018image,wu2019pseudo,wu2017global} of feature map from each channel in intermediate layers.

Moreover, You \etal \cite{multi_teachers} revealed that multiple teacher networks can provide more comprehensive knowledge for learning a more effective student network.  MEAL \cite{MEAL} compressed large and complex trained ensembles into a single network, which employs an adversarial-based learning strategy to guide the pre-defined student network to transfer knowledge from teacher models. Offline knowledge distillation requires a pre-trained teacher model in the stage of optimizing, while online KD simultaneously optimizes the teacher and student network or just a group of student peers \cite{KD_survey}. Anil \etal \cite{codistillation} trained two networks with the identical architecture parallelly and these two networks play the role of student and teacher iteratively.  
In this paper, we employ the multi-granularity based online distillation scheme, which aims to learn an effective student model from complementary structure of the teacher generators and the knowledge from diverse layers.

\section{Methodology}

In this section, we first introduce the proposed online GAN distillation framework, where the student generator is not bound to the discriminator and attempts to learn concepts directly from the teacher models.
The multi-granularity distillation scheme is presented in section \ref{MGD}
The multi-granularity concepts \cite{li2020multi} are captured via complementary teacher generators and knowledge from diverse layers. 
We illustrate the whole pipeline of the OMGD framework in Figure \ref{fig:MGOKD}.

\subsection{Online GAN Distillation}
Recently, a series of distillation based GAN compression \cite{GAN_compression,DMAD,AGD,portableGAN,TinyGAN,GAN_slimming} employ the offline distillation scheme that leverages a pre-trained teacher generator to optimize the student generator. 
In this paper, we propose a GAN-oriented online distillation algorithm to address three critical issues in the offline distillation.
Firstly, the student generator in the conventional offline distillation method should maintain a certain capacity to keep the dynamic balance with the discriminator to avoid model collapse \cite{peng2018variational, Salimans2016ImprovedTF} and vanishing gradients \cite{WGAN}. However, our student generator is no longer deeply bound with the discriminator, which can train more flexibly and obtain further compression.
Secondly, the pre-trained teacher generator fail to guide the student on how to learn information progressively and is easy to cause over-fitting in the training stage \cite{Guo_2020_CVPR, Lan2018KnowledgeDB}. Nevertheless, our teacher generator helps to warm up student generator and guide the direction of optimization step by step.
Thirdly, it is not effortless to select a suitable pre-trained teacher generator due to the evaluation metrics are subjective \cite{Salimans2016ImprovedTF}. However, our method does not require a pre-trained model and this selection problem is solved.

\textbf{Teacher Generator.} We follow the loss functions and training setting in \cite{pix2pix, cyclegan} to train the teacher generator $G_T$ and discriminator $D$. $G_T$ aims to learn a function to map data from the source domain $X$ to a target domain $Y$. 
We take Pix2Pix \cite{pix2pix} as example, it uses paired data($\{x_i,y_i\}_{i=1}^N$, where $x_i \in X$ and $y_i \in Y$) to optimize networks. The generator $G_T$ is trained to map $x_i$ to $y_i$ while the discriminator $D$ is trained to distinguish the fake images generated by $G_T$ from the real images. The objective is formalized as:  
\begin{equation}\label{eq:1}
\begin{split}
    \mathcal{L}_{GAN}(G_T,D)=&\mathbb{E}_{x,y}[\log D(x,y)]\\
                           &+\mathbb{E}_{x}[\log (1-D(x,G_T(x))].
\end{split}
\end{equation}

Moreover, a reconstruction loss is introduced to push the output of $G_T$ output to be close to the ground truth $y$:
\begin{equation}\label{eq:2}
    \mathcal{L}_{Recon}(G_T)=\mathbb{E}_{x,y}[\parallel{y-G_T(x)}\parallel_1].
\end{equation}

The whole objective in the GAN setting is defined as:
\begin{equation}\label{eq:3}
  G_T^*=\arg\min_{G_T}\max_{D}\mathcal{L}_{GAN}(G_T,D)+\mathcal{L}_{Recon}(G_T).
\end{equation}

\textbf{Student Generator.} In the proposed GAN-oriented online distillation scheme, the student generator $G_S$ only leverages the teacher network $G_T$ for optimization and  can be trained in the discriminator-free setting.
The optimization of $G_S$ does not require the ground-truth labels $y$ simultaneously. Namely, $G_S$ merely learns the output of the larger capacity generator with a similar structure ($G_T$), which greatly reduces the difficulty of fitting $y$ directly.
Specifically, we back-propagate the distillation loss between $G_T$ and $G_S$ in every iterative step. In this way, $G_S$ can mimic the training process of $G_T$ to learn progressively. 

Denote the output of $G_T/G_S$ as $p_t/p_s$, we use \emph{Structural Similarity (SSIM) Loss} \cite{SSIM} and \emph{Perceptual Losses} \cite{Perceptual} to measure the difference between $p_t$ and $p_s$. SSIM Loss \cite{SSIM} is sensitive to local structural changes, which is
similar to human visual system (HVS).
Given $p_s, p_t$, SSIM Loss calculates the similarity of two images by:
\begin{equation}\label{eq:4}
    \mathcal{L}_{SSIM}(p_t,p_s)=\frac{(2\mu_t\mu_s+C_1)(2\sigma_{ts}+C_2)}{(\mu_t^2\mu_s^2+C_1)(\sigma_t^2+\sigma_s^2+C_2)},
\end{equation}
where $\mu_s, \mu_t$ are mean values  for luminance estimation, $\sigma_s^2, \sigma_t^2$ are standard deviations for contrast, $\sigma_{ts}$ is covariance for the structural similarity estimation. $C_1, C_2$ are constants to avoid zero denominator.

Perceptual loss \cite{Perceptual} consists of feature reconstruction loss $\mathcal{L}_{feature}$ and style reconstruction loss $\mathcal{L}_{style}$. $\mathcal{L}_{feature}$ encourages $p_t$ and $p_s$ to have similar feature representations, which are measured by a pre-trained VGG network \cite{VGG} $\phi$. $\mathcal{L}_{feature}$ is formalized as: 
\begin{equation}\label{eq:5}
    \mathcal{L}_{feature}(p_t,p_s)=\frac{1}{C_j H_j W_j}\parallel{\phi_j(p_t)-\phi_j(p_s)}\parallel_1,
\end{equation}
where $\phi_j(x)$ is the activation of the $j$-th layer of $\phi$ for the input $x$. $C_j \times H_j \times W_j$ is the dimensions of $\phi_j(x)$.

$\mathcal{L}_{style}$ is introduced to penalize the differences in style characteristic, such as color, textures, common pattern, and so on \cite{Gatys2015ANA}. 
The $\mathcal{L}_{style}$ can be calculated as:
\begin{equation}\label{eq:7}
    \mathcal{L}_{style}(p_t,p_s)=\parallel{G_j^{\phi}(p_t)-G_j^{\phi}(p_s)}\parallel_1,
\end{equation}
where $G_j^{\phi}(x)$ is the \emph{Gram matrix}
of the $j$-th layer activation in the VGG network.

Furthmore, the total variation loss $\mathcal{L}_{TV}$ \cite{TV_Loss} is introdued to encourages spatial smoothness in the generated image. We use four hyperparameters $\lambda_{SSIM}, \lambda_{feature}, \lambda_{style}, \lambda_{TV}$ to achieve the balance between the above losses, so the total online KD loss $\mathcal{L}_{KD}(p_t,p_s)$ is computed by:
\begin{equation}\label{eq:8}
\begin{split}
    \mathcal{L}_{KD}(p_t,p_s)=&\lambda_{SSIM}\mathcal{L}_{SSIM}+\lambda_{feature}\mathcal{L}_{feature}\\&+\lambda_{style}\mathcal{L}_{style}+\lambda_{TV}\mathcal{L}_{TV}
\end{split}
\end{equation}

\subsection{Multi Granularity Distillation Scheme} \label{MGD}
Based on the novel online GAN distillation technique, we further extend our method into a multi-granularity scheme from two perspectives: the complementary structure of the teacher generator, and the knowledge from diverse layers. 
The whole pipeline of the online multi-granularity distillation (OMGD) framework is depicted in Figure \ref{fig:MGOKD}, we use a wider teacher generator $G_{T}^W$ and a deeper teacher generator $G_{T}^D$ to formalize a multi-objective optimization task for $G_S$. 
In addition to the output layer of the teacher generators, we also mine knowledge concepts from the intermediate layers via channel distillation loss \cite{channel_KD}.

\textbf{Multiple Teachers Distillation.} 
A different structure based teacher generator helps to capture more complementary image cues from the ground truth labels and enhance image translation performance from a different perspective \cite{multi_teachers}. 
Moreover, the multiple teachers distillation setting can further relieve the issue of over-fitting. 
We expand the student model into the teacher model from two complementary dimensions, i.e., depth and width. Given a student generator $G_{S}$, we expand the channel of $G_{S}$ to obtain a wider teacher generator $G_{T}^W$. Specifically, each channel of the convolution layers is multiplied by an channel expand factor $\eta$. 
On the other hand, we insert several resnet blocks after every downsample and upsample layers into $G_{S}$ to build a deeper teacher generator $G_{T}^D$, which has a comparable capacity with $G_{T}^W$. 

As is illustrated in Figure \ref{fig:MGOKD}, a partial share discriminator is designed to share the first several layers and separate two branches to get the discriminator output for $G_{T}^W$ and $G_{T}^D$, respectively.
This shared design not only offers high flexibility of discriminators but also leverages the similar characteristic of the input image to improve the training of generators \cite{slimmable_GAN}.
We directly combine two distillation losses provides by the complementary teacher generators as the KD loss in multiple teachers setting:
\begin{equation}\label{eq:9}
\begin{split}
        &\mathcal{L}_{KD_{multi}}(p_{t}^w,p_{t}^d,p_s)\\
        &=\mathcal{L}_{KD}(p_{t}^w,p_s)+\mathcal{L}_{KD}(p_{t}^d,p_s),
\end{split}
\end{equation}
where $p_{t}^w$ and $p_{t}^d$ are the activation of the output layer of $G_{T}^W$ and $G_{T}^D$, respectively.

\textbf{Intermediate Layers Distillation.} 
The concepts of the output layer fail to take into account of more intermediate details of the teacher network, so we further transfer the channel-wise granularity information as an additional supervisory signal to promote $G_S$.
Specifically, we compute the channel-wise attention weight \cite{SENet,channel_KD} to measure the importance of each channel in a feature map. The attention weight $w_c$ is defined as:
\begin{equation}\label{eq:10}
     w_c = \frac{1}{H \times W}\sum_{i=1}^H\sum_{j=1}^Wu_c(i,j),
\end{equation}
where $u_c$ denotes the $c^{th}$ channel of the feature map. Then a $1 \times 1$ convolution layer is concatenated to the intermediate layers of $G_S$ to expand the number of channel and the channel distillation (CD) loss is calculated as:
\begin{equation}\label{eq:11}
    \mathcal{L}_{CD}(G_{T}^{W}, G_S)=\frac{1}{n}\sum_{i=1}^n(\frac{\sum_{j=1}^c(w_{t_w}^{ij}-w_{s}^{ij})^2}{c}),
\end{equation}
where $n$ is the number of feature maps be sampled, $c$ is the channel number of the feature maps. $w^{ij}$ is the attention weight of $j$-th channel of $i$-{th} feature map.

To sum up, the whole online multi-granularity distillation objective is formalized as:
\begin{equation}\label{eq:12}
    \begin{split}
         &\mathcal{L}(G_{T}^W, G_{T}^D, G_S)\\
         &=\lambda_{CD}\mathcal{L}_{CD}(G_{T}^W, G_S)+\mathcal{L}_{KD_{multi}}(p_t^w,p_{t}^d,p_s)
    \end{split}
    \end{equation}



\section{Experiments} \label{sec:experiments}
\subsection{Experimental Settings}

\begin{table*}[ht]
    \centering
    \caption{The performance comparison with state-of-the-art methods in Pix2Pix model.}
    \vspace{-5pt}
    \begin{tabular}{m{2.0cm}<{\centering}|m{2.4cm}<{\centering}|m{3.6cm}<{\centering}|m{2.8cm}<{\centering}|m{2.8cm}<{\centering}|m{1.5cm}<{\centering}}
    \toprule
   Dataset & Generator Style & Method &  MACs & \#Parameters & FID ($\downarrow$)\\\hline
     \multirow{10}{*}{edges$\rightarrow$shoes}  
     & \multirow{5}{*}{Res-Net} &Original \cite{pix2pix} & 56.80G (1.0$\times$)& 11.30M (1.0$\times$) &24.18  \\
    & & GAN-Compression \cite{GAN_compression} & 4.81G (11.8$\times$)& 0.70M (16.3$\times$) &26.60 \\
    &	& DMAD \cite{DMAD} & 4.30G (13.2$\times$)& 0.54M (20.9$\times$) &24.08 \\ \cline{3-6}
    &	& OMGD 1.0$\times$ & 1.408G (40.3$\times$) & 0.137M (82.5$\times$) &25.88 \\ 
    &	& OMGD 1.5$\times$ & \textbf{2.904G (19.6$\times$)} & \textbf{0.296M (38.2$\times$)} &\textbf{21.41} \\
    \cline{2-6}
    & \multirow{5}{*}{U-Net}&	Original \cite{pix2pix} & 
    18.60G (1.0$\times$) & 54.40M (1.0$\times$) & 34.31  \\
    & &DMAD \cite{DMAD} & 2.99G (6.2$\times$) & 2.13M (25.5$\times$) & 46.95  \\ \cline{3-6}
    & &OMGD 0.5$\times$ & 0.333G (55.9$\times$) & 0.852M (63.8$\times$) & 37.34  \\ 
    & &OMGD 0.75$\times$ & 0.707G  (26.3$\times$) & 1.916M (28.4$\times$) & 32.19  \\ 
    & &OMGD 1.0$\times$ & \textbf{1.219G (15.3$\times$)} & \textbf{3.404M (16.0$\times$)} &\textbf{25.00}  \\ 
    \midrule
    Dataset & Generator Style & Method &  MACs & \#Parameters &mIoU ($\uparrow$)\\\hline
    \multirow{11}{*}{cityscapes}  
    & \multirow{6}{*}{Res-Net} & Original \cite{pix2pix} &56.80G (1.0$\times$) & 11.30M (1.0$\times$)  &44.32 \\
    &	&GAN-Compression \cite{GAN_compression} & 5.66G (10$\times$) & 0.71M (15.9$\times$)  &40.77 \\
    &	&DMAD \cite{DMAD} & 4.39G (12.9$\times$) & 0.55M (20.5$\times$)  &41.47 \\
    &	&CAT \cite{CAT} & 5.57G (10.2$\times$) & -  & 42.53 \\ \cline{3-6}
    &	&OMGD 1.0$\times$& 1.408G (40.3$\times$) &  0.137M (82.5$\times$)   & 45.21 \\
    &	&OMGD 1.5$\times$& \textbf{2.904G (19.6$\times$)} & \textbf{0.296M (38.2$\times$)}   & \textbf{45.89} \\
    \cline{2-6}
    & \multirow{5}{*}{U-Net}&	Original \cite{pix2pix} & 
    18.60G (1.0$\times$) & 54.40M (1.0$\times$) & 42.71  \\
    &	&DMAD \cite{DMAD} & 3.96G (4.7$\times$) & 1.73M (31.4$\times$) &40.53 \\ \cline{3-6}
    & & OMGD 0.5$\times$ & 0.333G (55.9$\times$) & 0.852M (63.8$\times$) & 41.54 \\ 
    & & OMGD 0.75$\times$ & 0.707G  (26.3$\times$) & 1.916M (28.4$\times$) & 45.52 \\ 
    & & OMGD 1.0$\times$ & \textbf{1.219G (15.3$\times$)} & \textbf{3.404M (16.0$\times$)} &\textbf{48.91} \\ 
    \bottomrule
    \end{tabular}
    \vspace{-5pt}
    \label{table:result1} 
  \end{table*}

We follow the models, datasets and evaluation metrics used in previous works \cite{AGD, GAN_compression,DMAD,co_evolution} for a fair comparison.

\textbf{Models.} We conduct the experiments on Pix2Pix \cite{pix2pix} and CycleGAN \cite{cyclegan}. Specifically, we adopt the original U-Net style generator \cite{pix2pix} and the Res-Net style generator in \cite{GAN_compression} for Pix2Pix \cite{pix2pix} model. The Res-Net style generator employs depthwise convolution and pointwise convolution \cite{MobileNet_V1} to achieve a better performance-computation trade-off. We only use the Res-Net style generator \cite{GAN_compression} to conduct the experiments on CycleGAN model.

\textbf{Datasets and Evaluation Metrics.} We evaluate Pix2Pix model on edges$\rightarrow$shoes \cite{Yu2014FineGrainedVC} and cityscapes \cite{Cordts2016TheCD} dataset. CycleGAN model is measured on horse$\rightarrow$zebra \cite{cyclegan} and summer$\rightarrow$winter \cite{cyclegan}. On cityscapes, we use the DRN-D-105 \cite{DRN} to segment the generated images and calculate the mIoU (mean Intersection over Union) as evaluation metric. Higher mIoU implies the generated images are more realistic. We adopt FID (Frechet Inception Distance) \cite{FID} to evaluate the image on other datasets and smaller FID means the generation performance is more convincing.

\textbf{Implementation datails.} 
The channel expand factor $\eta$ is set to 4 in this paper. We set the learning rate as 0.0002 in the beginning and decay to zero linearly in the experiments. 
For Res-Net style generator, batch size is set to 4 on edges$\rightarrow$shoes and 1 on other dataset. Batch size is fixed to 4 in all expreiments for U-Net generator. The update interval numbers $n$ on edges$\rightarrow$shoes, cityscapes, horse$\rightarrow$zebra and summer$\rightarrow$winter are 1, 3, 4, 4 respectively. 

\begin{table*}[ht]
    \renewcommand\arraystretch{1.0}
    \centering
    \caption{The performance comparison with state-of-the-art methods in CycleGAN model.}
    \vspace{-5pt}
    \begin{tabular}{m{2.8cm}<{\centering}|m{3.8cm}<{\centering}|m{2.5cm}<{\centering}|m{2.5cm}<{\centering}|m{1.8cm}<{\centering}}
    \toprule
    Dataset  & Method &  MACs & \#Parameters & FID ($\downarrow$)\\ \hline
    \multirow{8}{*}{horse$\rightarrow$zebra}  
    & Original \cite{cyclegan} & 56.80G (1.0$\times$) & 11.30M  (1.0$\times$) & 61.53 \\
    &	Co-Evolution \cite{co_evolution} & 13.40G (4.2$\times$)  &- & 96.15 \\
    &	GAN-Slimming \cite{GAN_slimming}  & 11.25G (23.6$\times$)  & - & 86.09 \\
    &	Auto-GAN-Distiller \cite{AGD}  & 6.39G (8.9$\times$) &- &83.60 \\
    &	GAN-Compression \cite{GAN_compression}  & 2.67G (21.3$\times$) & 0.34M (33.2$\times$) & 64.95 \\
    &	DMAD \cite{DMAD}  & 2.41G (23.6$\times$)  & 0.28M (40.0$\times$) & 62.96 \\
    &	CAT \cite{CAT}  & 2.55G (22.3$\times$)  & - & 60.18 \\
    &	OMGD (Ours) & \textbf{1.408G (40.3$\times$)} & \textbf{0.137M (82.5$\times$)}  & \textbf{51.92} \\\hline
    \multirow{5}{*}{summer$\rightarrow$winter}  
    & Original \cite{cyclegan} & 56.80G (1.0$\times$) & 11.30M  (1.0$\times$) & 79.12  \\
    &	Co-Evolution \cite{co_evolution} & 11.10G (5.1$\times$)  & - &78.58 \\
    &	Auto-GAN-Distiller \cite{AGD}  & 4.34G (13.1$\times$) &- &78.33 \\
    &	DMAD \cite{DMAD}  & 3.18G (17.9$\times$) & 0.30M (37.7$\times$) & 78.24 \\
    &	OMGD (Ours) & \textbf{1.408G (40.3$\times$)}  & \textbf{0.137M (82.5$\times$)} & \textbf{73.79} \\
    \bottomrule
    \end{tabular}
    \vspace{-5pt}
    \label{table:result2}
     \end{table*}

For CycleGAN, we evaluate the teacher generator every $m$ epochs and update the performance-best generator as $G_T$ to optimize $G_S$. 
In this way, we avoid notorious instability of training CycleGAN and encourage $G_S$ to learn from the best teacher model.
$m$ is set to 10 and 6 for horse$\rightarrow$zebra and summer$\rightarrow$winter, respectively.

\subsection{Experimental Results}
\subsubsection{Comparison with state-of-the-art methods} 

In this section, we compare OMGD with several state-of-the-art methods in terms of computation cost, model size and generation quality. We compare the performance of Pix2Pix and CycleGAN respectively.

\textbf{Pix2Pix.} The experimental results of Pix2Pix model are shown in Table \ref{table:result1}, which can summarized as the following observerations:
1) OMGD is robust for both style generators and significantly outperforms the state-of-the-art methods with much less computational costs.
2) OMGD with Res-Net style generator (dubbed as, OMGD(R)) achieves comparable performance to the original model when the MACs are compressed by \textbf{40.3$\times$} and the parameters are compressed by \textbf{82.5$\times$}.
Compared with the current best method, i.e., CAT, OMGD(R) 1.0$\times$ boosts the mIoU from 42.53 to 45.21 (6.3\% improvement) with only a quarter of the computational costs on cityscapes.
Furthmore, although OMGD(R) 1.5$\times$ is compressed \textbf{19.6$\times$} MACs and \textbf{38.2$\times$} memory,  it successes to establishes the state-of-the-art performance.
3) It is arduous to compress U-Net style generator due to its U-shape architecture and concatenate operation. OMGD with U-Net style generator (dubbed as, OMGD(U)) compresses the original model by \textbf{15.3$\times$} and reduces the FID by 9.31 on edges$\rightarrow$shoes. With less than half of MACs of DMAD \cite{DMAD},  OMGD(U) 1.0$\times$ decreases the FID from 46.95 to 25.0 on edges$\rightarrow$shoes  and obtains 19.3\% improvement in terms of mIoU on cityscapes. 
Moreover, OMGD(U) 0.5$\times$ and 0.75$\times$ also achieve impressive results, and OMGD(U) 0.75$\times$ can obatin the state-of-the-art compression performance with merely \textbf{0.707G} MACs.

\begin{table}[t]
    \renewcommand\arraystretch{1.0}
    \centering
    \caption{Ablation Study on Pix2Pix model.}
    \vspace{-5pt}
    \begin{tabular}{m{2.2cm}<{\centering}|m{3.0cm}<{\centering}|m{1.5cm}<{\centering}}
    \toprule
Dataset  & Method & FID ($\downarrow$)\\\hline
    \multirow{5}{*}{edges$\rightarrow$shoes}  
    & Baseline & 77.07 \\ 
    &	Ours w/o OD & 26.19 \\
    & Ours w/o DT & 33.88 \\
    &	Ours w/o CD & 26.62 \\ \cline{2-3}
    &	Ours & \textbf{25.00}  \\    
    \midrule
    Dataset  & Method & mIoU ($\uparrow$) \\ \hline
    \multirow{5}{*}{cityscapes}  
    & Baseline & 33.90 \\
    &	Ours w/o OD & 45.76 \\
    & Ours w/o DT  & 44.04 \\
    &	Ours w/o CD & 48.12 \\ \cline{2-3}
    &	Ours & \textbf{48.91} \\
    \bottomrule
    \end{tabular}
    \vspace{-5pt}
    \label{table:result3}
\end{table}

\textbf{CycleGAN.} 
We follow previous works \cite{GAN_compression,AGD,CAT,AutoGAN,GAN_slimming} to use the Res-Net style generator to conduct the experiments on CycleGAN, and the results are shown in Table \ref{table:result2}.
On the one hand, OMGD(R) outperforms the original model by a large margin although with \textbf{40.3$\times$} MACs compression and \textbf{82.5$\times$} parameters compression.
For example, OMGD(R) reduces FID from 61.53 to 51.92 on horse$\rightarrow$zebra and 79.12 to 73.79 on summer$\rightarrow$winter.
On the other hand, OMGD(R) significantly surpasses the competitive methods in terms of performance (FID) or computational costs (MACs), and establishes new state-of-the-art performance on both datasets.


\begin{table}[t]
    \renewcommand\arraystretch{1.0}
    \centering
    \caption{Ablation Study on CycleGAN model.}
    \vspace{-5pt}
    \begin{tabular}{m{2.5cm}<{\centering}|m{3.0cm}<{\centering}|m{1.5cm}<{\centering}}
    \toprule
    Dataset  & Method & FID ($\downarrow$)\\\hline
    \multirow{4}{*}{horse$\rightarrow$zebra}  
    & Baseline & 96.72 \\
    &	Ours w/o OD & 77.09 \\
    &	Ours w/o CD & 61.21 \\\cline{2-3}
    &	Ours  & \textbf{51.92} \\
    \midrule
    \multirow{4}{*}{summer$\rightarrow$winter}  
    & Baseline & 78.43 \\
    &	Ours w/o OD & 76.48 \\
    &	Ours w/o CD & 75.47 \\\cline{2-3}
    &	Ours  & \textbf{73.79}  \\
    \bottomrule
    \end{tabular}
    \vspace{-5pt}
    \label{table:result4}
    \end{table}

\begin{figure*}[ht]
    \centering
    \includegraphics[width=17.8cm,height=8.8cm]{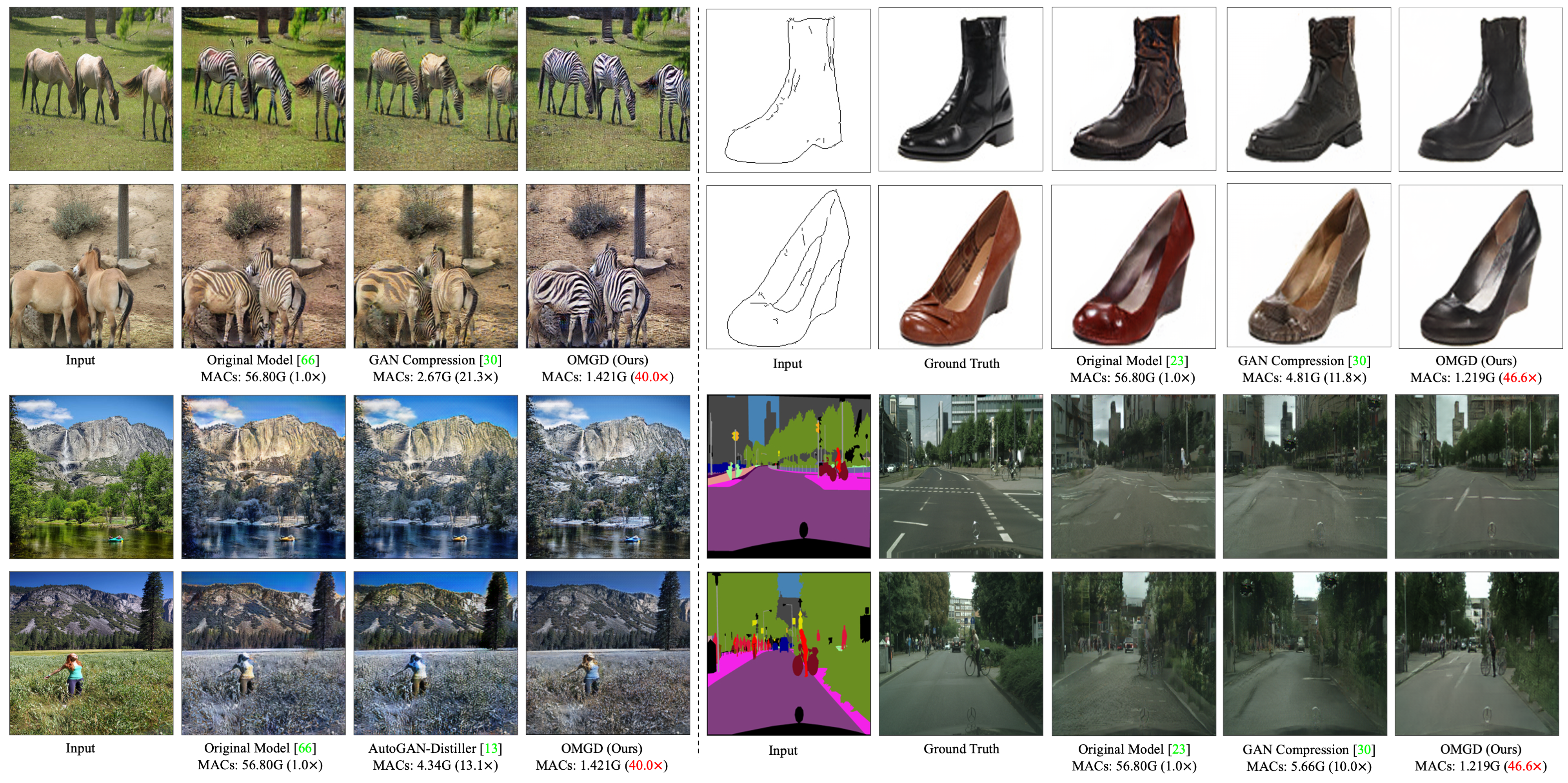}
    \vspace{-20pt}
    \caption{Qualitative compression results on horse$\rightarrow$zebra, summer$\rightarrow$winter, cityscapes and edges$\rightarrow$shoes. OMGD preserves the fidelity while significantly reducing the computation.}
    \vspace{-5pt}
    \label{fig:qualitative}
    \end{figure*}

\subsubsection{Ablation Study} 

We directly train the student generator via the conventional GAN loss and report its results as the “Baseline" in Table \ref{table:result3} and Table \ref{table:result4}. As can be observed, our method surpasses “Baseline" by a large margin.
For example, it declines FID from 77.07 to 25.00 on edges$\rightarrow$shoes and increases mIoU from 33.90 to 48.91 on cityscapes. 
To further demonstrate the effectiveness of several essential components in OMGD, we perform extensive ablation studies. 
The experiments of ablation study are conducted on U-Net style generator for Pix2Pix and Res-Net style generator for CycleGAN, 

\textbf{Analysis of online distillation stage.} 
To evaluate the significance of the online distillation scheme, we design a variant (abbreviated as “Ours w/o OD”) to optimize the model with the offline two-stage distillation setting.  
As shown in Table \ref{table:result3} and \ref{table:result4}, removing the online distillation stage leads to an noticeable drop in performance.
For example, “Ours w/o OD” declines mIoU to 45.76 on cityscapes, with a decrease of 6.4\% when compared with our approach. 
It indicates that the online training scheme helps to guide the optimization process to achieve more impressive results.
     
\textbf{Analysis of complementary teachers setting.} 
To investigate the effectiveness of complementary teachers setting, we design a variant “Ours w/o DT” that removes the deeper teacher generator and only employs a wider one for optimization. As summarized in Table \ref{table:result3}, our method attempts to obtain more promising results compared with “Ours w/o DT” on both benchmarks.   
It indicates that the complementary teacher setting significantly improves the capacity of the student generator. 
It is worth notice that the unstable training process of CycleGAN causes confusion for the deeper teacher generator $G_T^D$, hence we only leverage the wider teacher generator on CycleGAN. 
     
\textbf{Analysis of multiple distillation layers.} 
To delve deep into the significance of multiple distillation layers, we design a variant (denote as “Ours w/o CD”) to remove the channel-wise distillation.
As shown in Table \ref{table:result3} and Table \ref{table:result4}, “Ours w/o CD” 
gets a less prominent performance, which indicates that concepts from intermediate layers can be viewed as the auxiliary supervision to assist training. 
By introducing multiple distillation layers for distillation, our method manages to obtain 6.5\%, 1.6\%, 15.1\% and 1.5\% performance improvement on four datasets, respectively.

\begin{table}[t]
   \renewcommand\arraystretch{1.0}
   \small
   \centering
   \caption{Latency Speedup on Mobile Phones (CPU).}
   \vspace{-8pt}
\begin{tabular}{c|c|c|c}
\hline
\multirow{2}{*}{Device} & original & \multicolumn{2}{c}{OMGD(U) 1.0$\times$} \\ \cline{2-4} 
                       & Latency  & Latency         & MAC           \\ \hline
HuaWei P20              & 416.73ms & 43.00ms (9.7$\times$)    & 15.3 $\times$ \\ \hline
Mi 10                   & 140.80ms & 14.01ms (10.0 $\times$)  & 15.3 $\times$ \\ \hline
\end{tabular}
   \vspace{-10pt}
   \label{table:Latency}
\end{table}

\subsubsection{Latency Speedup}
We report the CPU latency results on two mobile phones (i.e., Huawei P20 and Mi 10) using tflite toolkits.
As is shown in Table \ref{table:Latency},  our framework helps to obtain significant acceleration in the inference procedure.
For example, OMGD(U) 1.0$\times$ contributes to reducing latency from 140.8ms to 14.01ms, with a \textbf{90\%} latency decline.
It demonstrates that our framework provides a solution for real-time image translation.

\subsubsection{Qualitative Results}
We depict the visualization results of OMGD and the state-of-the-art methods in Figure \ref{fig:qualitative}, which demonstrates the effectiveness of OMGD. 
As shown, our method helps to obtain 40.3-46.6 $\times$ MACs reductions with nearly no visual fidelity loss.
For example,  Our $40\times$ compressed model can generate natural zebra stripes on horse$\rightarrow$zebra dataset, while \cite{GAN_compression} and original model still retain the color of the input horse. 
OMGD attempts to transfer the background style smoothly, while preserves the essential elements in the foreground on summer$\rightarrow$winter.
For Pix2Pix, OMGD contributes to capturing the textural details of the cloth fabric and the shine of the leather fabric on edges$\rightarrow$shoes. Furthermore, OMGD shows superiority in the processing of pavement features, such as roughness and lane line.

\section{Conclusion}
In this paper, we propose an online multi-granularity distillation (OMGD) technique for learning lightweight GAN.
The GAN-oriented online scheme is introduced to alternately promote the teacher and student generator, and the teacher helps to warm up the student and guide the optimization direction step by step. 
OMGD further takes good advantage of multi-granularity concepts from complementary teacher generators and auxiliary supervision signals from different layers. 
Extensive experiments demonstrate that OMGD attempts to compress Pix2Pix and CycleGAN into extremely low computational costs without obvious visual fidelity loss, which 
provides a feasible solution for GAN deployment on resource-constrained devices.

{\small
\bibliographystyle{ieee_fullname}
\bibliography{egbib}
}

\clearpage

\section{Appendix}

\subsection{Additional Implementation Details}

\begin{algorithm}[h]
    \caption{The Learn-Best Strategy for CycleGAN.}
    \label{alg:1}
    \begin{algorithmic}[1]
    \State \textbf{Input:} training set $\mathcal{D}_{train}$, validation set $\mathcal{D}_{val}$, CycleGAN generators $G_A$, $G_B$, discriminators $D$, student generator $G_S$, update interval $n$, evaluate interval $m$
    \State \textbf{Output:} trained student generator $G_S^*$
    \State Initialize $G_A$, $G_B$, $D$, $G_S$, $F_A^{best}=\infty$ , $F_B^{best}=\infty$ 
    \For {$epochs=1,...,T$}
        \For {$steps=1,...,K$}
            \State Sample a mini-batch data: $x=sample(\mathcal{D})$;
            \State Calculate $\mathcal{L}_{CycleGAN}(G_A, G_B,D)$;
            \State Update parameters of $G_A$, $G_B$, $D$;
        \EndFor
        \If{$epochs$ $ \% $ $m$ == 0}
            \State Calculate FID $F_A$ of $G_A$ on $\mathcal{D}_{val}$;
            \State Calculate FID $F_B$ of $G_B$ on $\mathcal{D}_{val}$;
            \If{($F_A<F_A^{best}$) or ($F_B<F_B^{best}$)}
                \State $F_A^{best}$ = $F_A$ or $F_B^{best}$ = $F_B$;
                \State Update $G_{T}^W$ by $G_{T}^W$ = $G_A$;
            \EndIf
            \For {$epochs=1,...,m \times n$}
                \State Sample a mini-batch data: $x=sample(\mathcal{D})$;
                \State $p_{t}^w=G_{T}^W(x)$; $p_s=G_S(x)$;
                \State Calculate $\mathcal{L}_{KD}(p_{t}^w,p_s)$ by Eq.7;
                \State Calculate $\mathcal{L}_{CD}(G_{T}^W, G_S)$ by Eq.10;
                \State Update parameters of $G_S$ by Eq.11;
            \EndFor
        \EndIf
    \EndFor    
    \end{algorithmic}
\end{algorithm}

\textbf{Learn-best strategy for CycleGAN.} To relieve the distillation difficulties brought by the notoriously unstable training process of CycleGAN, we developed a simple yet effective strategy in the training stage. Specifically, we evaluate the symmetrical generators in CycleGAN every $m$ epochs, then employ the best one as the teacher generator to refine the student generator, the whole distillation process of CycleGAN with the learn-best strategy is illustrated in Algorithm \ref{alg:1}.

\begin{figure}[ht]
    \centering
    \includegraphics[width=0.55\linewidth]{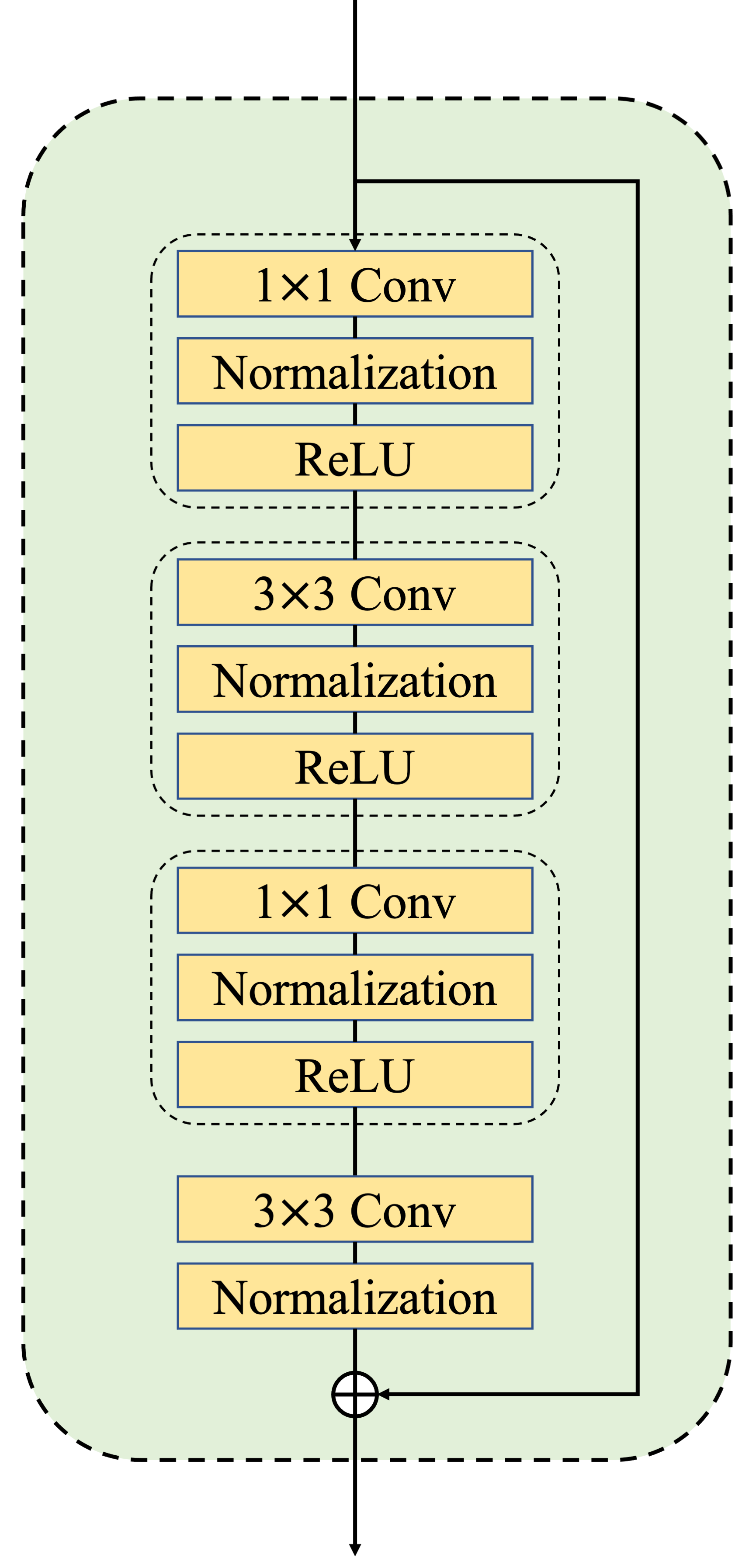}
    \vspace{-5pt}
    \caption{Resnet Block in deeper teacher generator.}
    \vspace{-5pt}
    \label{fig:res_block}
\end{figure}

\textbf{Deeper teacher generator.} 
Figure \ref{fig:res_block} illustrates the construction of the resnet block from the deeper teacher generator. For the U-Net style generator, we insert two resnet blocks after every downsample and upsample layer to obtain the deeper teacher generator. For the Res-Net style generator, we insert a resnet block and an additional 3$\times$3 convolution layer after every downsample and upsample layer. Furthermore, we also add two resnet blocks in the middle of the student generator's backbone to construct the deeper teacher generator.

\begin{table*}[ht]
    \centering
    \caption{Hyper-parameters setting in OMGD.}
    \vspace{-10pt}
    \begin{tabular}{m{2.2cm}<{\centering}|m{2.1cm}<{\centering}|m{0.9cm}<{\centering}|m{0.9cm}<{\centering}|m{0.8cm}<{\centering}|m{0.8cm}<{\centering}|m{0.8cm}<{\centering}|m{0.8cm}<{\centering}|m{0.8cm}<{\centering}|m{0.8cm}<{\centering}|m{0.8cm}<{\centering}|m{0.8cm}<{\centering}}
    \toprule
    \multirow{2}{*}{Dataset} & \multirow{2}{*}{Generator Style} & \multicolumn{2}{c|}{Training Epochs} & \multirow{2}{*}{$n$} & \multirow{2}{*}{$m$} & \multicolumn{3}{c|}{ngf} & \multirow{2}{*}{ndf} & \multirow{2}{*}{$\lambda_{CD}$} & \multirow{2}{*}{D$_{share}$} \\  \cline{3-4} \cline{7-9}
     & & Const& Decay & &   & $G_T^W$ & $G_T^D$ & $G_S$ & &  &                             \\ \hline
    \multirow{2}{*}{edges$\rightarrow$shoes} 
    & Res-Net &100  & 100 & 1 & - & 64 & 16 & 16 & 128 & 1e2 & 3 \\
    & U-Net  & 100 &  100  & 1 & - & 64 & 16  & 16  & 128 & 1e1 & 5 \\ \hline
    \multirow{2}{*}{cityscapes}& 
    Res-Net & 300 & 450 & 3 & - & 64 & 16 & 16 & 128 & 0 & 4 \\
    & U-Net & 300 & 450 & 3 & - & 64 & 16 & 16 & 128 & 5e1 & 5 \\ \hline
    horse$\rightarrow$zebra 
    & Res-Net &  100 &  100 & 4 &  10  & 64  & - & 16 & 64 &5e2 & - \\ \hline
    summer$\rightarrow$winter 
    & Res-Net  & 100  & 100  & 4 & 6  & 64  & -   & 16 & 64 &1e2  & - \\
    \bottomrule
    \end{tabular}
    \vspace{5pt}
    \label{table:hyper-parameters}
    \end{table*}

\textbf{Hyper-parameters setting.} 
$\lambda_{SSIM}$, $\lambda_{feature}$, $\lambda_{style}$, $\lambda_{TV}$ are set as 1e1,1e4,1e1,1e-5, respectively. 
Following the previous works \cite{cyclegan, pix2pix, GAN_compression}, we apply Adam optimizer \cite{Adam} in all experiments. 
More details of the hyper-parameters setting are shown in Table \ref{table:hyper-parameters}. “Training Epochs” means the training epochs for the student generator, “Const” is the epochs that keeps the fixed initial learning rate and “Decay” is the epochs of linearly decaying learning rate. $n$ denotes the update interval for the teacher generator. For example, we train the student generator for 750 epochs on cityscapes and set the update interval to 3, which means that we train the teacher generator for 250 epochs. $m$ is the evaluate interval in our learn-best strategy, “ngf” and “ndf” denote the base number of filters in generator and discriminator respectively, which control the model size. $\lambda_{CD}$ means the weight of channel distillation loss and $D_{share}$ controls the number of shared layers in our partial shared discriminator. In the experiment, we found that channel distillation loss can not show a significant effect for the Res-Net style generator on the cityscapes dataset, so we set $\lambda_{CD}$ to zero.
For the teacher networks, we use the same setting in \cite{GAN_compression}, Hinge GAN loss \cite{GeometricGAN, Self-Attention-GAN} is employed to Pix2Pix and LSGAN loss \cite{LSGAN} is employed to CycleGAN.

\subsection{Additional Ablation Study}

In this section, we further investigate several important components of our method. Experiments are conducted on U-Net style generator for Pix2Pix and Res-Net style generator for CycleGAN.

\begin{table}[t]
    \centering
    \caption{Ablation study of discriminator-free setting and the learn-best strategy.}
    \vspace{-5pt}
    \begin{tabular}{m{2.3cm}<{\centering}|m{2.4cm}<{\centering}|m{0.9cm}<{\centering}m{0.9cm}<{\centering}}
    \toprule
    Dataset                                    & Method          & FID($\downarrow$) & mIoU($\uparrow$) \\ \hline
    \multirow{2}{*}{edges$\rightarrow$shoes}   
    & Ours w/ $L_{GAN}$ &  73.44  & -  \\
    & Ours                  & \textbf{25.00}  & -  \\ \hline
    \multirow{2}{*}{cityscapes} 
    & Ours w/ $L_{GAN}$ &   -   &  31.79 \\
    & Ours                  &  -   &  \textbf{48.91} \\ \hline
    \multirow{3}{*}{horse$\rightarrow$zebra}   
    & Ours w/ $L_{GAN}$ &  65.67 & -  \\
    & Ours w/o LB  & 63.15 \\
    & Ours                  & \textbf{52.00}  &  -\\ \hline
    \multirow{3}{*}{summer$\rightarrow$winter} 
    & Ours w/ $L_{GAN}$ &  75.53  &  - \\
    & Ours w/o LB & 75.70 \\
    & Ours                  &  \textbf{74.36} &  - \\
    \bottomrule
    \end{tabular}
    \vspace{-5pt}
    \label{table:gan_best}
    \end{table}

\textbf{Analysis of discriminator-free setting.} To measure the significance of the discriminator-free setting for the student generator, we design a variant that introduces a discriminator and employ GAN loss $L_{GAN}$ for training (denote as “Ours w/ $L_{GAN}$”). Table \ref{table:gan_best} shows that Ours w/ $L_{GAN}$ gets worse results, which reveals that the unstable optimization process and the invalid concepts from the discriminator influence the performance of the student generator. 
For example, the discriminator setting increases FID from 25.00 to 73.44 on edges$\rightarrow$shoes and declines mIoU from 48.91 to 31.79 on Pix2Pix model.

\textbf{Analysis of the learn-best strategy.}  To verify the effectiveness of our learn-best strategy on CycleGAN, we introduce a variant (abbreviated as “Ours w/o LB”)  that removes the learn-best strategy and distill student under the basic OMGD scheme.
Table \ref{table:gan_best} shows that learn-best strategy contributes to declining FID from 63.15 to 52.00 on horse$\rightarrow$zebra and from 75.70 to 74.36 on summer$\rightarrow$winter.
It further indicates that the learn-best strategy is capable of alleviating the impact of teacher's training instability.

\begin{table}[t]
    \centering
    \caption{Ablation Study of Architecture Complementarity.}
    \vspace{-5pt}
    \begin{tabular}{m{1.8cm}<{\centering}|m{1.4cm}<{\centering}|m{1.4cm}<{\centering}|m{0.9cm}<{\centering}m{0.9cm}<{\centering}}
    \toprule
    \multirow{2}{*}{Dataset} & \multicolumn{2}{c|}{Architecture} & \multirow{2}{*}{FID($\downarrow$)} & \multirow{2}{*}{mIoU($\uparrow$)} \\ \cline{2-3}
    & Teacher 1  & Teacher 2  & &                                   \\ \hline
    \multirow{3}{*}{edges$\rightarrow$shoes} 
    & W  & W  & 34.28 & - \\
    & D   & D  & 26.00 & - \\ 
    & W  & D  & \textbf{25.00}  & -   \\ \hline
    \multirow{3}{*}{cityscapes}              
    &W  & W &-  & 46.39  \\
    &D  & D   & - & 46.65 \\ 
    &W & D  &  - & \textbf{48.91} \\
    \bottomrule
    \end{tabular}
    \vspace{-5pt}
     \label{table:same_teachers}
    \end{table}

\begin{table*}[ht]
    \renewcommand\arraystretch{1.0}
    \centering
    \caption{The performance of different size model.}
    \vspace{-5pt}
    \begin{tabular}{m{2.8cm}<{\centering}|m{3.8cm}<{\centering}|m{2.5cm}<{\centering}|m{2.5cm}<{\centering}|m{1.8cm}<{\centering}|m{1.8cm}<{\centering}}
    \toprule
    Dataset  & Method &  MACs & \#Parameters & FID ($\downarrow$) & mIoU ($\uparrow$)\\ \hline
     \multirow{6}{*}{edges$\rightarrow$shoes}
    &	OMGD (U) 0.5$\times$ & 0.333G & 0.852M & 37.34 & -\\
    &	OMGD (U) 0.75$\times$ & 0.707G & 1.916M & 32.30 &- \\
    &	OMGD (U) 1.0$\times$ & 1.219G  & 3.404M & 25.00 &- \\ \cline{2-6}
    &	OMGD (R) 0.5$\times$ & 0.446G  & 0.039M & 38.06 & - \\
    &	OMGD (R) 0.75$\times$ & 0.867G & 0.081M & 34.48 &- \\
    &	OMGD (R) 1.0$\times$ & 1.421G & 0.137M & 25.88 & -\\\hline
     \multirow{6}{*}{cityscapes}
    &	OMGD (U) 0.5$\times$ & 0.333G & 0.852M & - & 41.54 \\
    &	OMGD (U) 0.75$\times$ & 0.707G & 1.916M  & - & 45.52 \\
    &	OMGD (U) 1.0$\times$ & 1.219G  & 3.404M  & - & 48.91 \\ \cline{2-6}
    &	OMGD (R) 0.5$\times$ & 0.446G  & 0.039M & - & 37.65 \\
    &	OMGD (R) 0.75$\times$ & 0.867G & 0.081M & - &  42.15 \\
    &	OMGD (R) 1.0$\times$ & 1.421G & 0.137M  & - & 45.21 \\\hline
    \multirow{3}{*}{horse$\rightarrow$zebra}  
    &	OMGD (R) 0.5$\times$ & 0.446G  & 0.039M & 71.27  & - \\
    &	OMGD (R) 0.75$\times$ & 0.867G & 0.081M  & 64.25 &- \\
    &	OMGD (R) 1.0$\times$ & 1.421G & 0.137M  & 52.00 &- \\\hline
    \multirow{3}{*}{summer$\rightarrow$winter}  
    &	OMGD (R) 0.5$\times$ & 0.446G  & 0.039M  & 75.46 & -\\
    &	OMGD (R) 0.75$\times$ & 0.867G & 0.081M & 74.95 & -\\
    &	OMGD (R) 1.0$\times$ & 1.421G & 0.137M & 74.36 &- \\
    \bottomrule
    \end{tabular}
    \vspace{-10pt}
    \label{table:result}
\end{table*}

\textbf{Analysis of the complementarity of multiple teachers.} A wider teacher generator and a deeper teacher generator helps to maintain a complementarity in the structure dimension,  which is very critical to break through the bottleneck of capacity for models with low computational costs. To further verify this motivation, we use two identical teacher generators to distill the student generator. As is shown in Table \ref{table:same_teachers} (“W” denotes the wider teacher generator, “D” denotes the deeper teacher generator), the complementary structure declines FID from 34.28 to 25.00 on edges$\rightarrow$shoes and increases mIoU from 46.39 to 48.91 on cityscapes.

\textbf{Analysis of further compression.} To comprehensively show the capability of OMGD, we set the “ngf” as 8 and 12 to obtain further compressed networks of both Pix2Pix and CycleGAN on four datasets (i.e., edges$\rightarrow$shoes, cityscapes, horse$\rightarrow$zebra, summer$\rightarrow$winter). Results about MACs, parameters as well as quantitative evaluation metrics of image fidelity are listed in Table \ref{table:result}. OMGD (R) means the Res-Net style generator and OMGD (U) denotes the  U-Net generator.
Table \ref{table:result} reveals that there is still some compression space for OMGD.
Furthermore, OMGD (U) 0.5$\times$ (only requires \textbf{0.333G MACs}) successes to achieve impressive results, which provides a feasible solution for deployment on resource-constrained devices and even breaks the barriers to real-time image translation on mobile devices.

\subsection{Additional Qualitative Results}

\begin{figure*}[ht]
    \centering
    \includegraphics[width=0.85\linewidth]{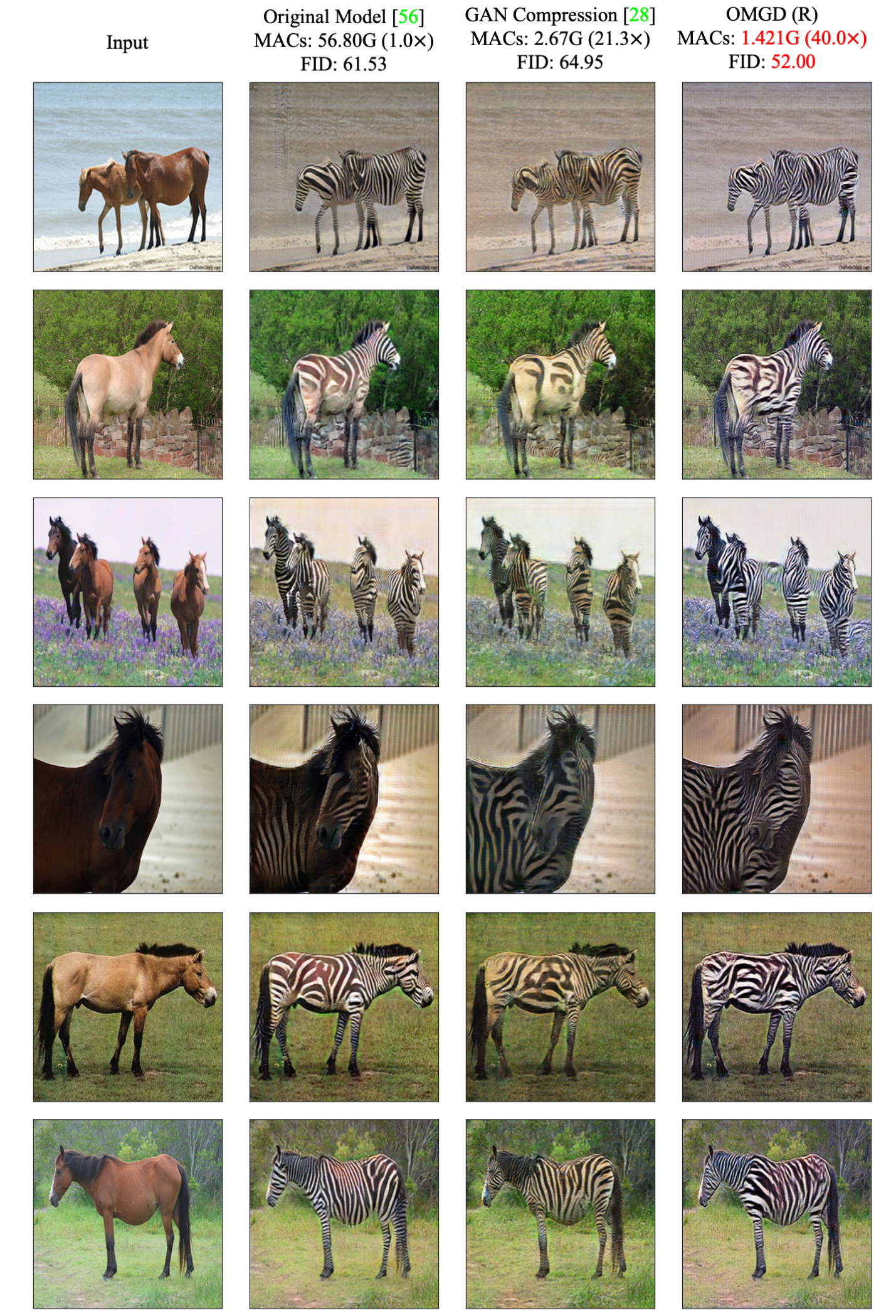}
    \vspace{-5pt}
    \caption{Additional qualitative results of OMGD with comparison to the state-of-the-art method and the original model on horse$\rightarrow$zebra dataset. OMGD (R) denotes that the Res-Net style generator is used.}
    \vspace{-5pt}
    \label{fig:vis1}
\end{figure*}

\begin{figure*}[ht]
    \centering
    \includegraphics[width=0.85\linewidth]{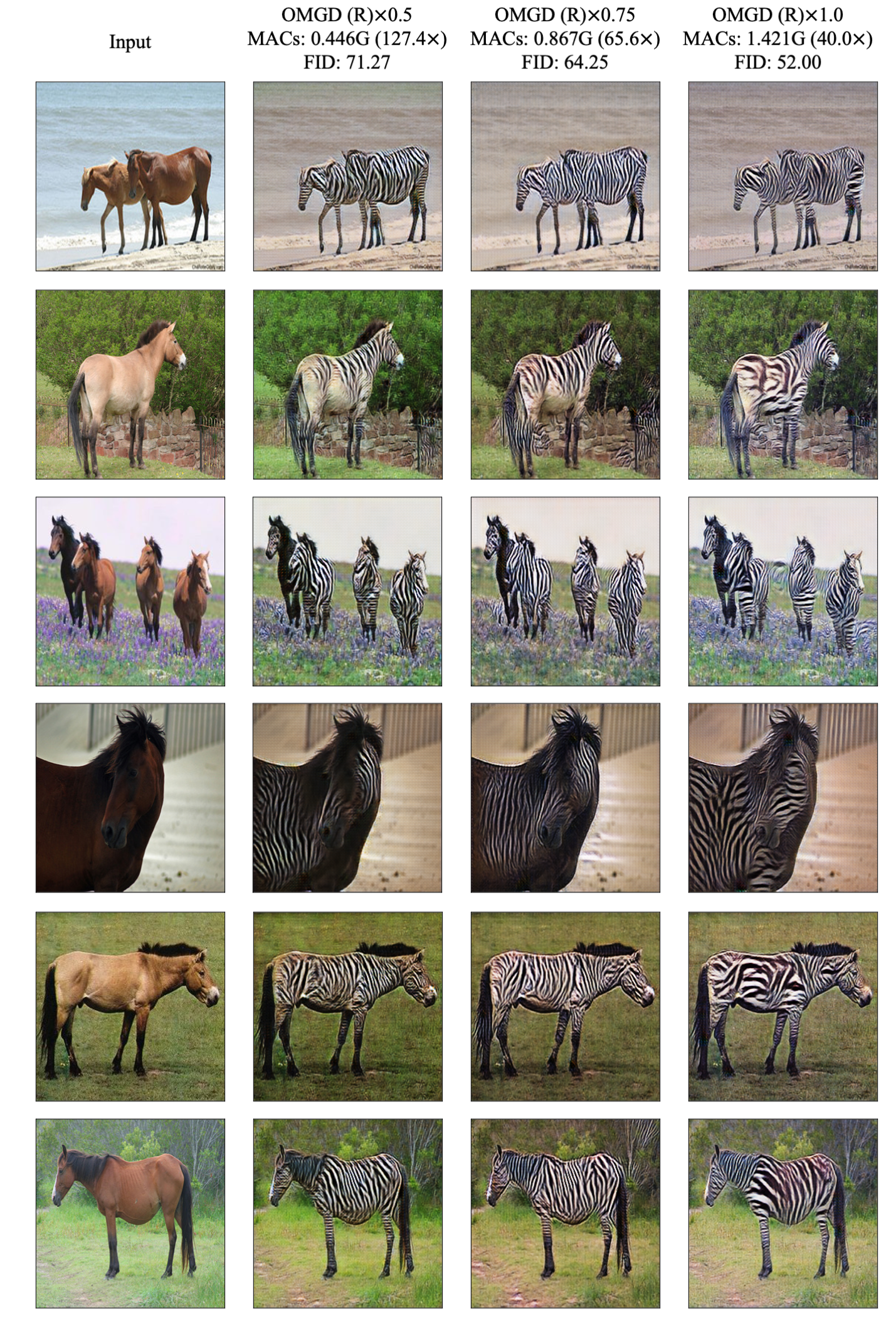}
    \vspace{-5pt}
    \caption{Qualitative results of different size models on horse$\rightarrow$zebra dataset. OMGD (R) denotes that the Res-Net style generator is used.}
    \vspace{-5pt}
    \label{fig:vis2}
\end{figure*}

\begin{figure*}[ht]
    \centering
    \includegraphics[width=0.85\linewidth]{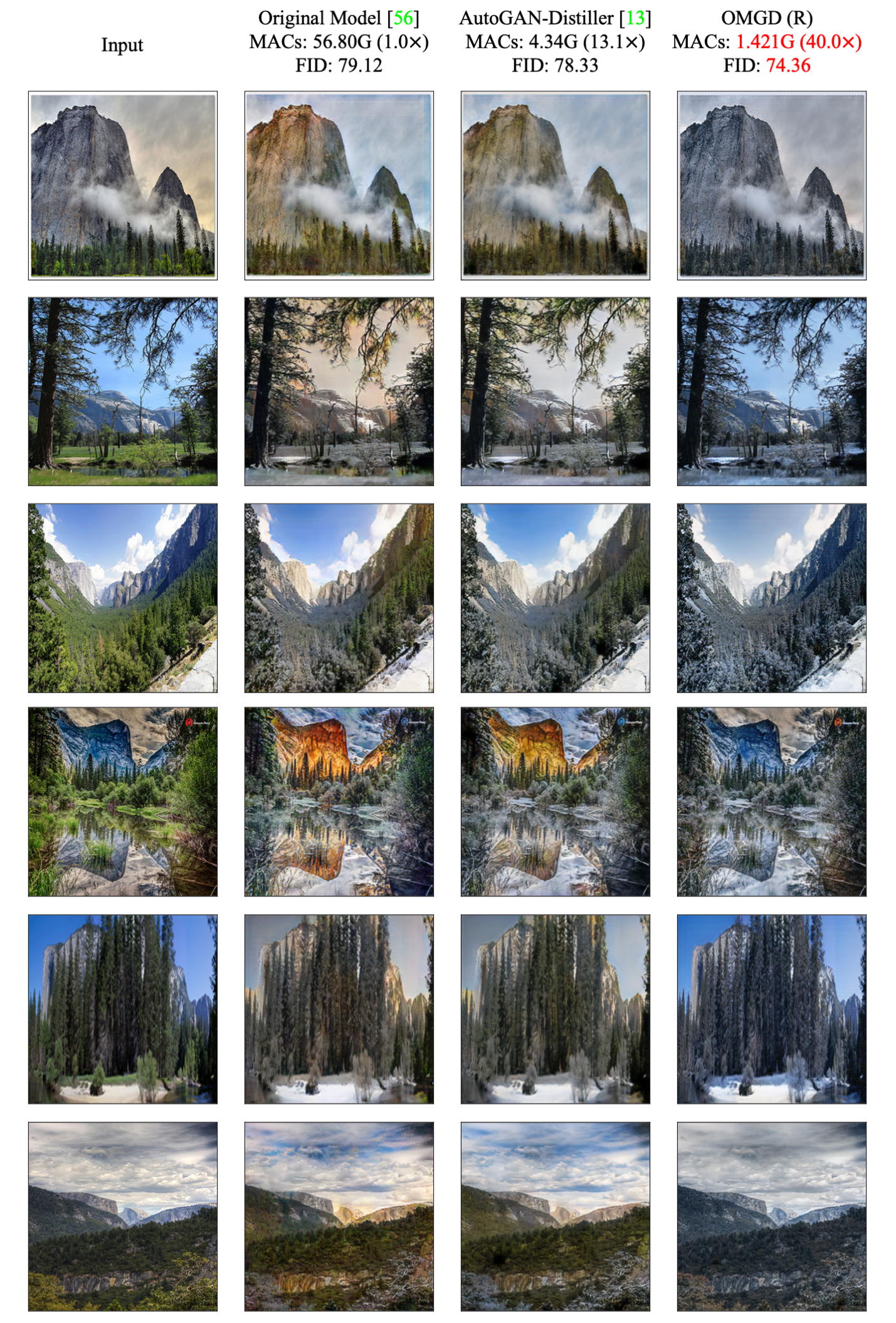}
    \vspace{-5pt}
    \caption{Additional qualitative results of OMGD with comparison to the state-of-the-art method and the original model on summer$\rightarrow$winter dataset. OMGD (R) denotes that the Res-Net style generator is used.}
    \vspace{-5pt}
    \label{fig:vis3}
\end{figure*}

\begin{figure*}[ht]
    \centering
    \includegraphics[width=0.9\linewidth]{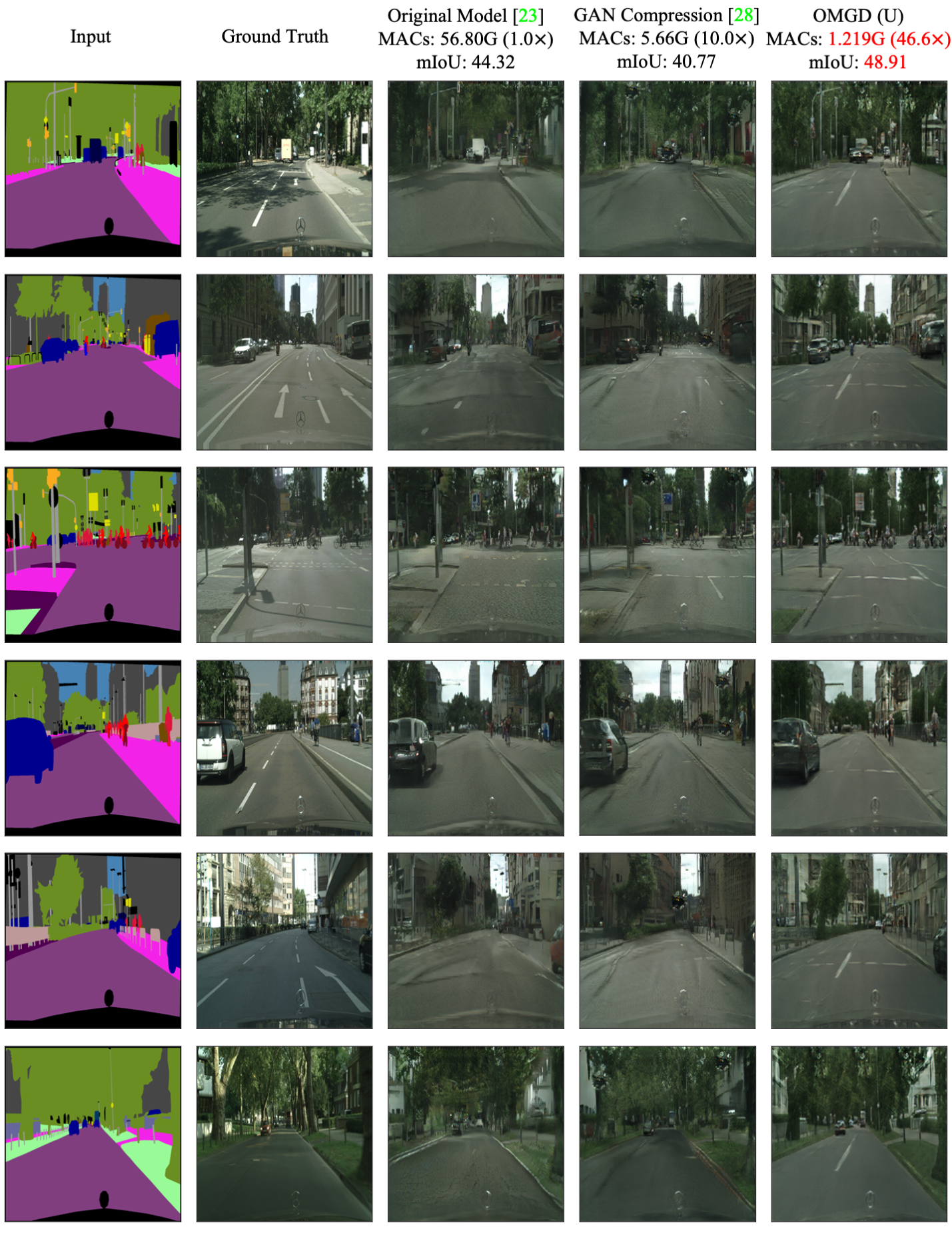}
    \vspace{-5pt}
    \caption{Additional qualitative results of OMGD with comparison to the state-of-the-art method and the original model on cityscapes dataset. OMGD (U) denotes that the U-Net style generator is used. }
    \vspace{-5pt}
    \label{fig:vis5}
\end{figure*}

\begin{figure*}[ht]
    \centering
    \includegraphics[width=0.9\linewidth]{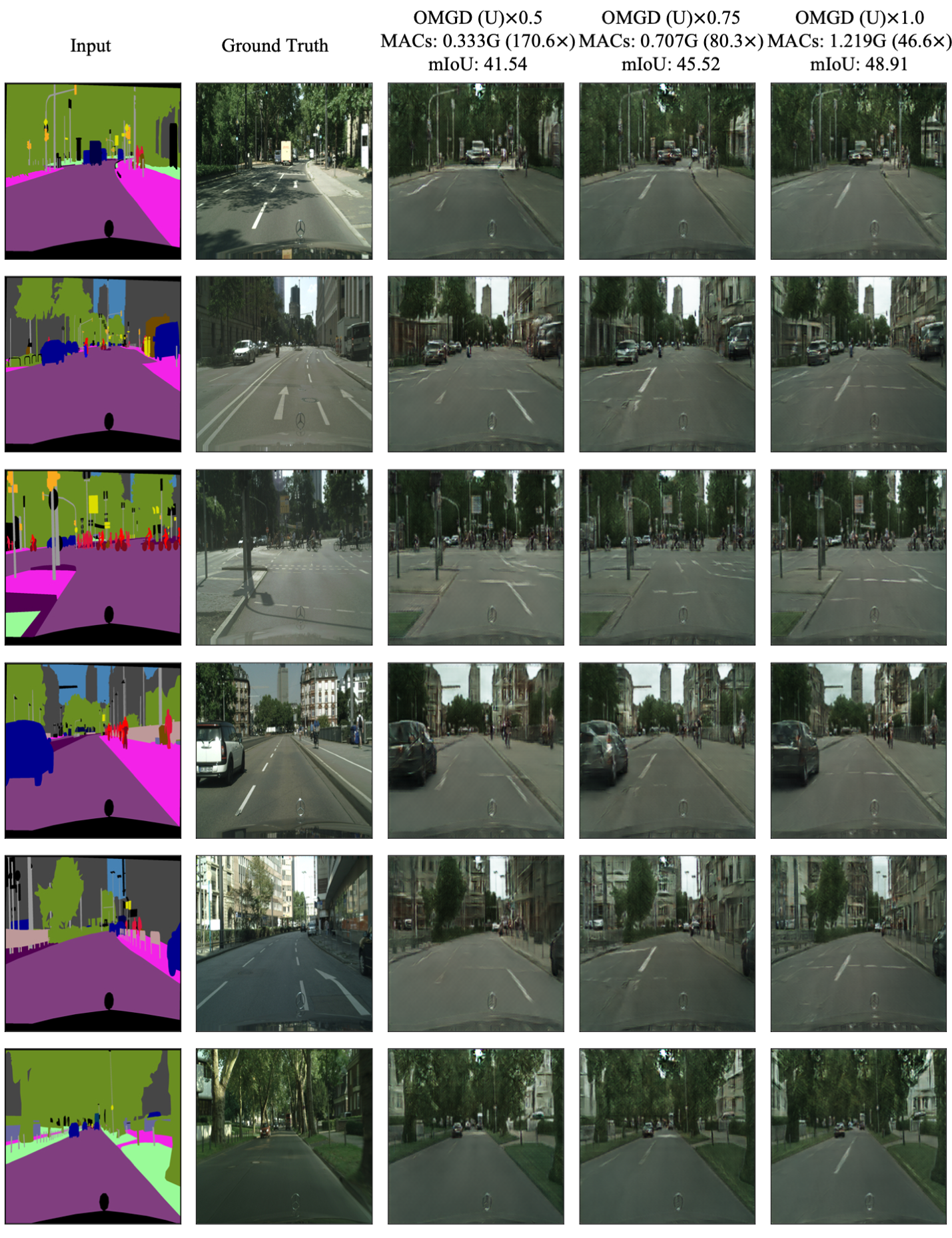}
    \vspace{-5pt}
    \caption{Qualitative results of different size models on cityscapes dataset. OMGD (U) denotes that the U-Net style generator is used.}
    \vspace{-5pt}
    \label{fig:vis6}
\end{figure*}

\begin{figure*}[ht]
    \centering
    \includegraphics[width=0.9\linewidth]{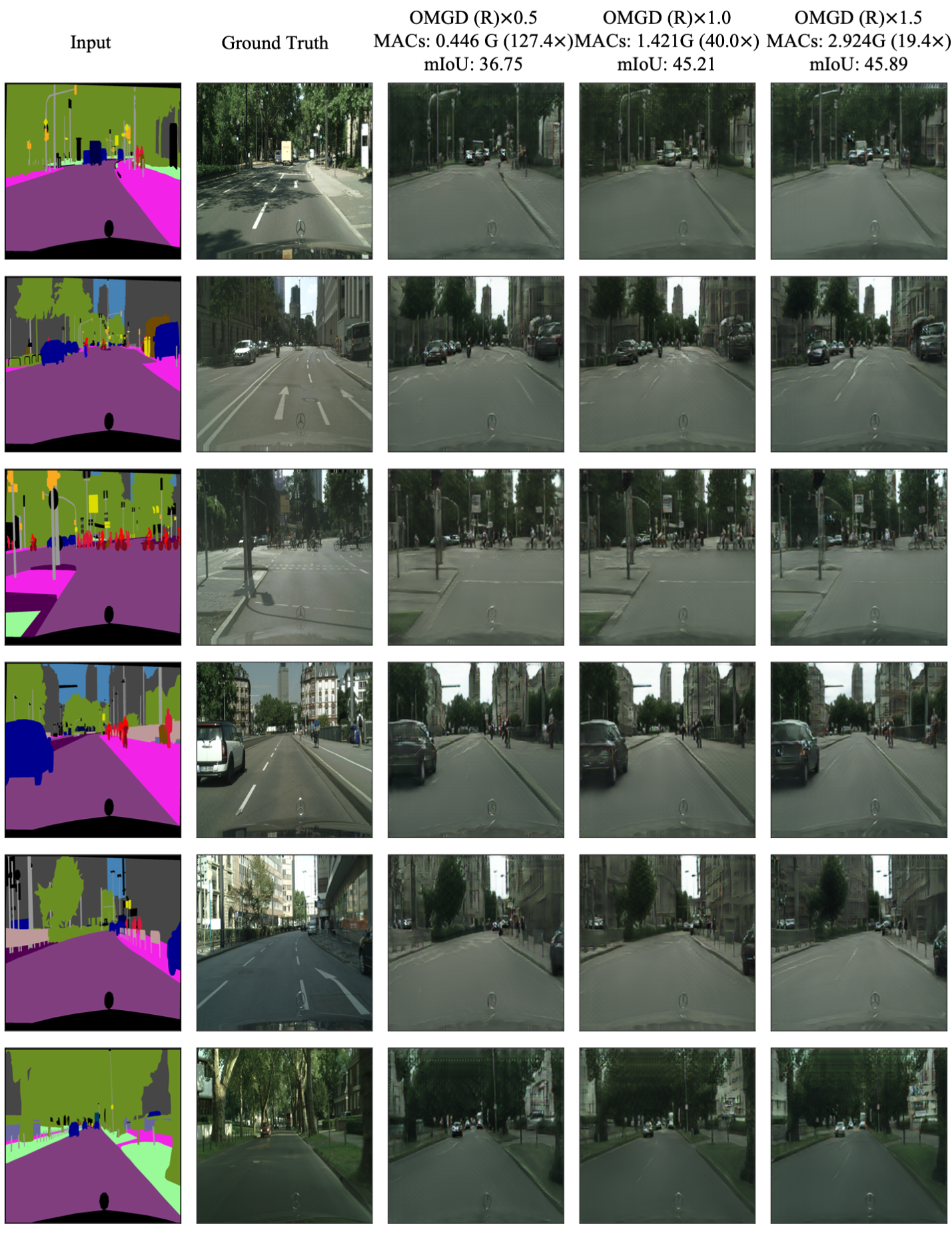}
    \vspace{-5pt}
    \caption{Qualitative results of different size models on cityscapes dataset. OMGD (R) denotes that the Res-Net style generator is used.}
    \vspace{-5pt}
    \label{fig:vis7}
\end{figure*}

\begin{figure*}[ht]
    \centering
    \includegraphics[width=0.9\linewidth]{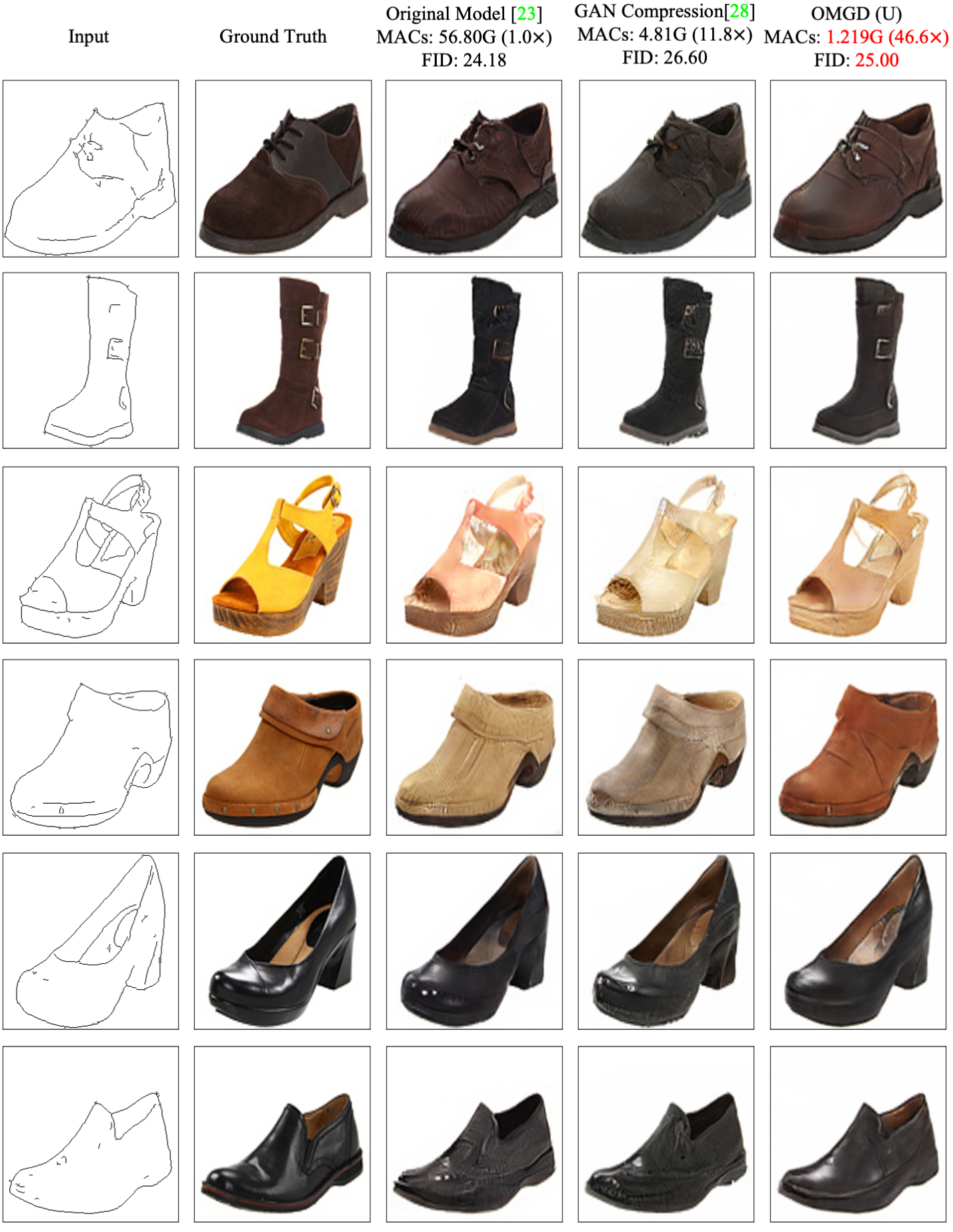}
    \vspace{-5pt}
    \caption{Additional qualitative results of OMGD with comparison to the state-of-the-art method and the original model on edges$\rightarrow$shoes dataset. OMGD (U) denotes that the U-Net style generator is used. }
    \vspace{-5pt}
    \label{fig:vis8}
\end{figure*}

\begin{figure*}[ht]
    \centering
    \includegraphics[width=0.9\linewidth]{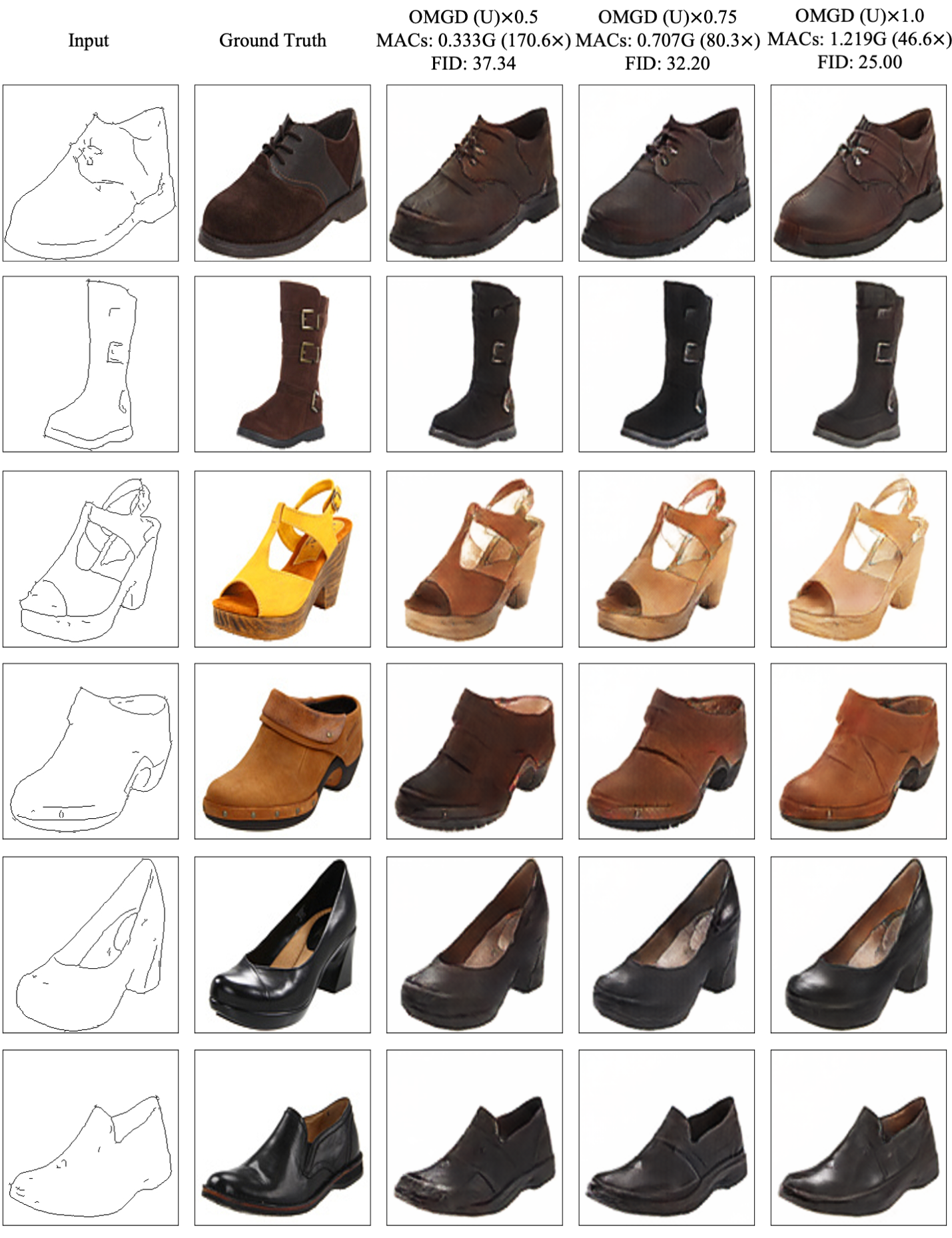}
    \vspace{-5pt}
    \caption{Qualitative results of different size models on edges$\rightarrow$shoes dataset. OMGD (U) denotes that the U-Net style generator is used.}
    \vspace{-5pt}
    \label{fig:vis9}
\end{figure*}

\begin{figure*}[ht]
    \centering
    \includegraphics[width=0.9\linewidth]{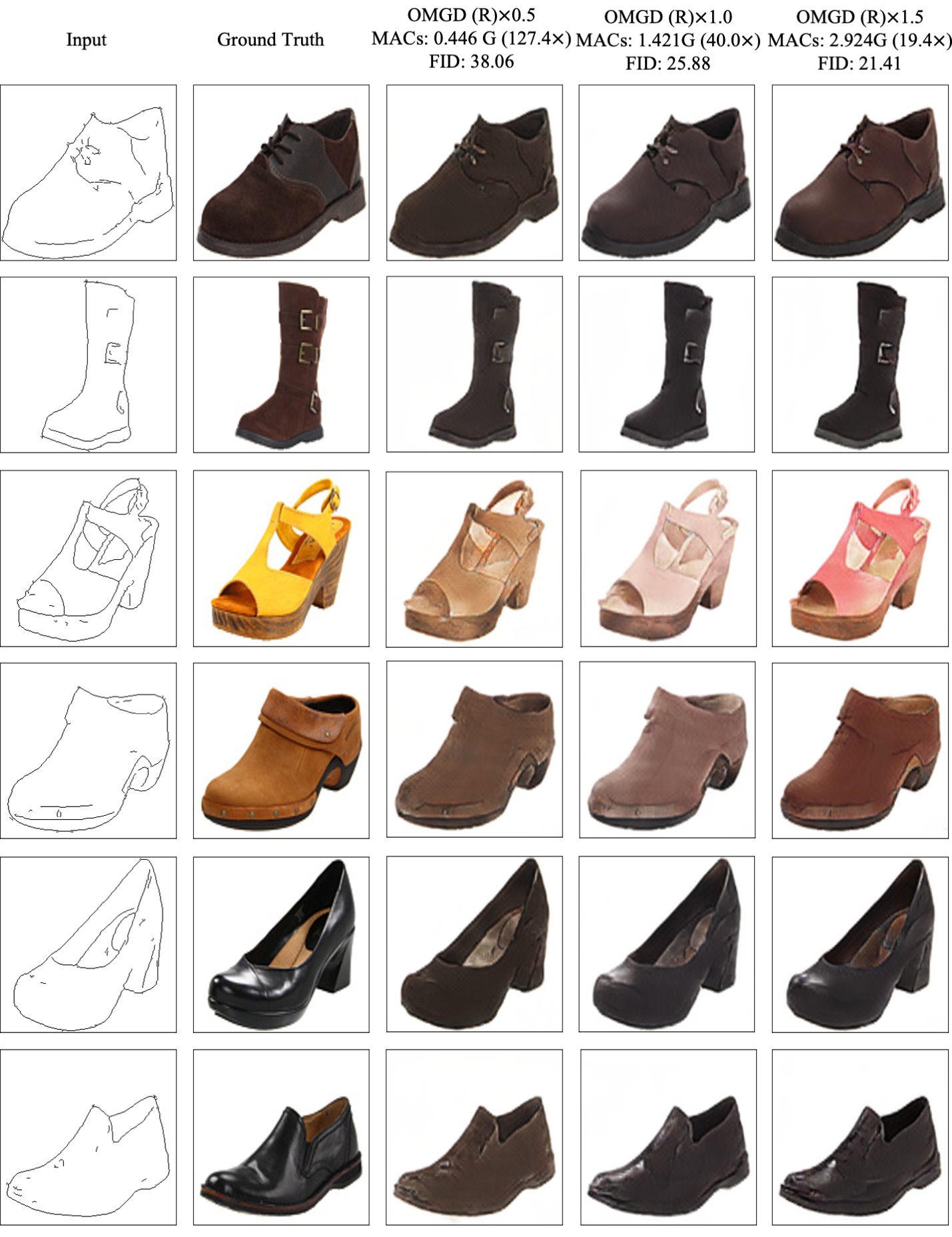}
    \vspace{-5pt}
    \caption{Qualitative results of different size models on edges$\rightarrow$shoes dataset. OMGD (R) denotes that the Res-Net style generator is used.}
    \vspace{-5pt}
    \label{fig:vis10}
\end{figure*}

We additionally provide more visualization results of OMGD and the state-of-the-art methods in Figure \ref{fig:vis1}, \ref{fig:vis3}, \ref{fig:vis5} and \ref{fig:vis8}, which demonstrates the effectiveness of OMGD. 
We also show the visualization results of different compression rate of OMGD method in Figure \ref{fig:vis2}, \ref{fig:vis6}, \ref{fig:vis7}, \ref{fig:vis9} and \ref{fig:vis10}.

\begin{figure*}[ht]
    \centering
    \includegraphics[width=0.85\linewidth]{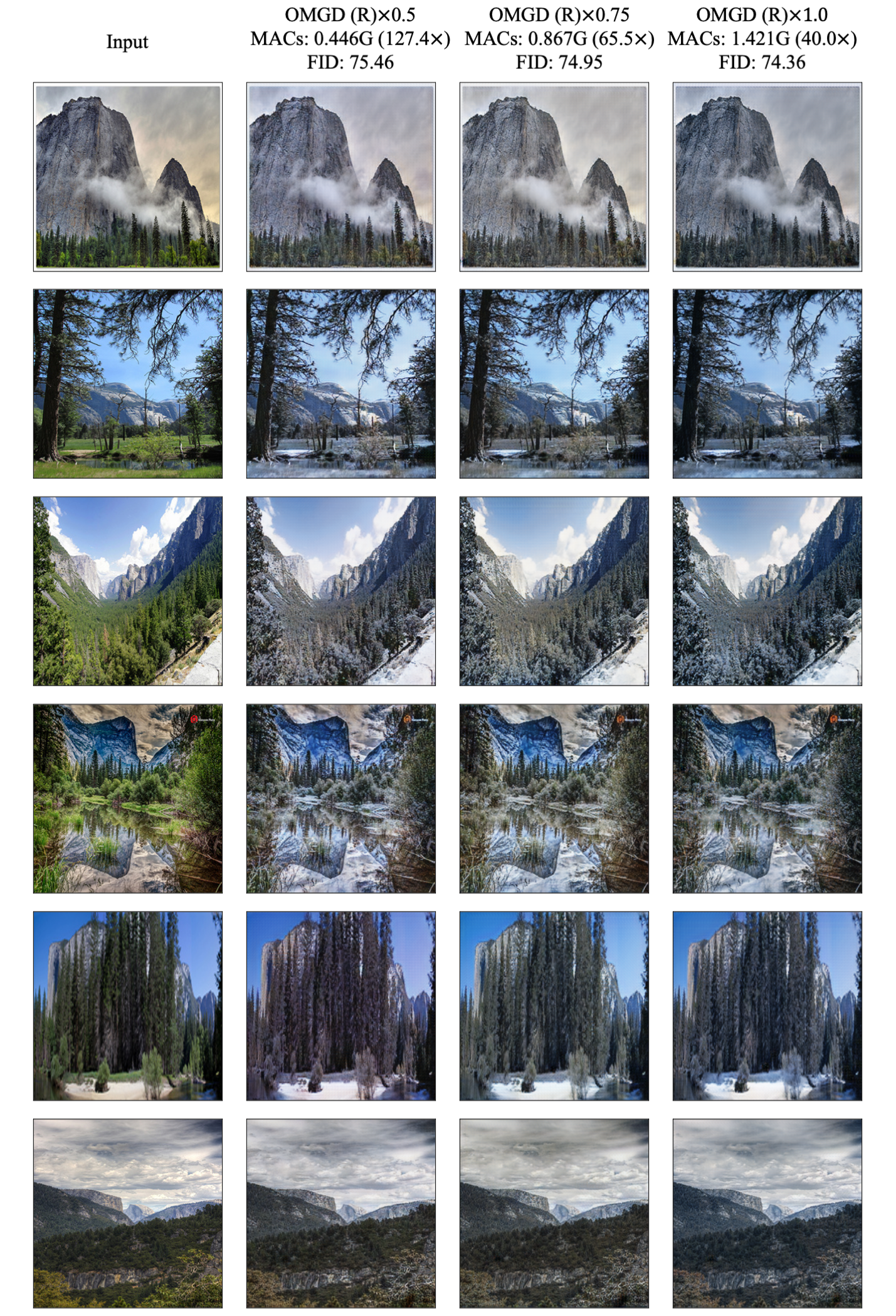}
    \vspace{-5pt}
    \caption{Qualitative results of different size models on summer$\rightarrow$winter dataset. OMGD (R) denotes that the Res-Net style generator is used.}
    \vspace{-5pt}
    \label{fig:vis4}
\end{figure*}

\end{document}